\documentclass[10pt,journal,compsoc]{IEEEtran}

\usepackage{mathrsfs}

\ifCLASSOPTIONcompsoc
  \usepackage[nocompress]{cite}
\else
  \usepackage{cite}
\fi

\usepackage{mathrsfs}
\usepackage[mathscr]{eucal}
\usepackage{amsmath}
\usepackage{array}
\usepackage{mathtools}
\usepackage{bbold}
\usepackage{bm}
\usepackage{booktabs}
\usepackage{epigraph}
\usepackage{cuted} 
\usepackage{lipsum}
\usepackage{comment}
\usepackage{pifont}

\usepackage{multirow}
\usepackage{xurl}

\usepackage[colorlinks=true, linkcolor=blue, citecolor=blue, urlcolor=blue]{hyperref}

\usepackage[sort, numbers]{natbib}
\setcitestyle{square}

\newcommand{\scalbijection}{h}
\newcommand{\actfunction}{h}
\newcommand{\obsvari}{Y}
\newcommand{\obsvar}{\mathbf{\obsvari}}
\newcommand{\obsvali}{y}
\newcommand{\obsval}{\mathbf{\obsvali}}
\newcommand{\obsset}{\mathcal{\obsvari}}
\newcommand{\obsdensity}{p_\obsvar}
\newcommand{\obsdata}{\mathcal{D}}
\newcommand{\obsnum}{M}
\newcommand{\couplingfunc}{\mathbf{\scalbijection}}
\newcommand{\couplingfunci}{\scalbijection}
\newcommand{\flowvari}{X}

\newcommand{\flowvali}{x}
\newcommand{\flowval}{\mathbf{\flowvali}}
\newcommand{\flowdim}{D}
\newcommand{\flownum}{N}
\newcommand{\flowfunc}{\mathbf{g}}
\newcommand{\flowinv}{\mathbf{f}}
\newcommand{\jac}[1]{\textrm{D}#1}
\newcommand{\flowfuncJ}{{\jac{\flowfunc}}}
\newcommand{\flowinvJ}{{\jac{\flowinv}}}
\newcommand{\flowparams}{\theta}
\newcommand{\basevari}{Z}
\newcommand{\basevar}{\mathbf{\basevari}}
\newcommand{\basevali}{z}
\newcommand{\baseval}{\mathbf{\basevali}}
\newcommand{\baseset}{\mathcal{\basevari}}
\newcommand{\basedensity}{p_\basevar}
\newcommand{\baseparams}{\phi}
\newcommand{\R}{\mathbb{R}}
\newcommand{\C}{\mathbb{C}}
\newcommand{\E}{\mathbb{E}}
\newcommand{\bigO}{\mathcal{O}}

\newcommand{\diag}{\textrm{diag}}
\newcommand{\tr}{\textrm{Tr}}

\newcommand{\KL}{D_{KL} }
\newcommand{\placeholder}{\boldsymbol{\cdot} }
\newcommand{\NN}{\text{NN} }
\newcommand{\eg}{\emph{e.g.}}
\newcommand{\ie}{\emph{i.e.}}

\newcommand{\BlackBox}{\rule{1.5ex}{1.5ex}}  

\usepackage{xcolor}
\usepackage{tocbibind}

\begin{document}

\title{Diffusion Models and Representation Learning: A Survey}
%
%
%
%

\author{Michael Fuest,~\IEEEmembership{}
  Pingchuan Ma,~\IEEEmembership{}
  Ming Gui,~\IEEEmembership{}
   Johannes  Schusterbauer,~\IEEEmembership{}
        Vincent Tao Hu,~\IEEEmembership{}
        Björn Ommer~\IEEEmembership{}
\IEEEcompsocitemizethanks{\IEEEcompsocthanksitem 
Michael Fuest is a Master's student at the Technical University of Munich. Pingchuan Ma, Ming Gui, and Johannes Schusterbauer are PhD students at LMU Munich. Vincent Tao Hu, a PostDoc from LMU Munich, is also the corresponding author.\protect\\
E-mail: taohu620@gmail.com
\IEEEcompsocthanksitem Björn Ommer is a full professor at LMU where he heads the Computer Vision \& Learning Group (previously Computer Vision Group Heidelberg).}
}

%
%

\markboth{IEEE Transactions on Pattern Analysis and Machine Intelligence}%
{Fuest \MakeLowercase{\textit{et al.}}: Diffusion Models and Representation Learning: A Survey}

%



\IEEEtitleabstractindextext{%
\begin{abstract}
Diffusion Models are popular generative modeling methods in various vision tasks, attracting significant attention. They can be considered a unique instance of self-supervised learning methods due to their independence from label annotation. This survey explores the interplay between diffusion models and representation learning.
It provides an overview of diffusion models' essential aspects, including mathematical foundations, popular denoising network architectures, and guidance methods. Various approaches related to diffusion models and representation learning are detailed. These include frameworks that leverage representations learned from pre-trained diffusion models for subsequent recognition tasks and methods that utilize advancements in representation and self-supervised learning to enhance diffusion models.
This survey aims to offer a comprehensive overview of the taxonomy between diffusion models and representation learning, identifying key areas of existing concerns and potential exploration. Github link: \url{https://github.com/dongzhuoyao/Diffusion-Representation-Learning-Survey-Taxonomy}.
\end{abstract}

}

\maketitle


\IEEEdisplaynontitleabstractindextext

%
\IEEEpeerreviewmaketitle


%
%
%
%


\IEEEraisesectionheading{\section{Introduction}\label{sec:introduction}}

Diffusion Models~\cite{sohl2015deep,song2021scorebased_sde,ho2020denoising} have recently emerged as the state-of-the-art of generative modeling, demonstrating remarkable results in image synthesis \cite{ho2020denoising,ho2021classifier,dhariwal_diffusion_2021,saharia_photorealistic_2022} and across other modalities including natural language \cite{austin_structured_2021, hoogeboom_argmax_2021, li_diffusion-lm_2022,flowseq}, computational chemistry \cite{anand_protein_2022,hoogeboom_equivariant_2022} and audio synthesis \cite{kong_diffwave_2021,liu_diffsinger_2022, huang_make--audio_2023}. The remarkable generative capabilities of Diffusion Models suggest that Diffusion Models learn both low and high-level features of their input data, potentially making them well-suited for general representation learning. Unlike other generative models like Generative Adversarial Networks (GANs) \cite{goodfellow_generative_2014,karras_style-based_2019,brock2018large_biggan} and Variational Autoencoders (VAEs) \cite{kingma2013auto_vae,rezende2014stochastic}, diffusion models do not contain fixed architectural components that capture data representations \cite{mittal_diffusion_2023}. This makes diffusion model-based representation learning challenging. Nevertheless, approaches leveraging diffusion models for representation learning have seen increasing interest, simultaneously driven by advancements in training and sampling of Diffusion Models.

\begin{figure}[t]
    \centering
    \includegraphics[width=\columnwidth]{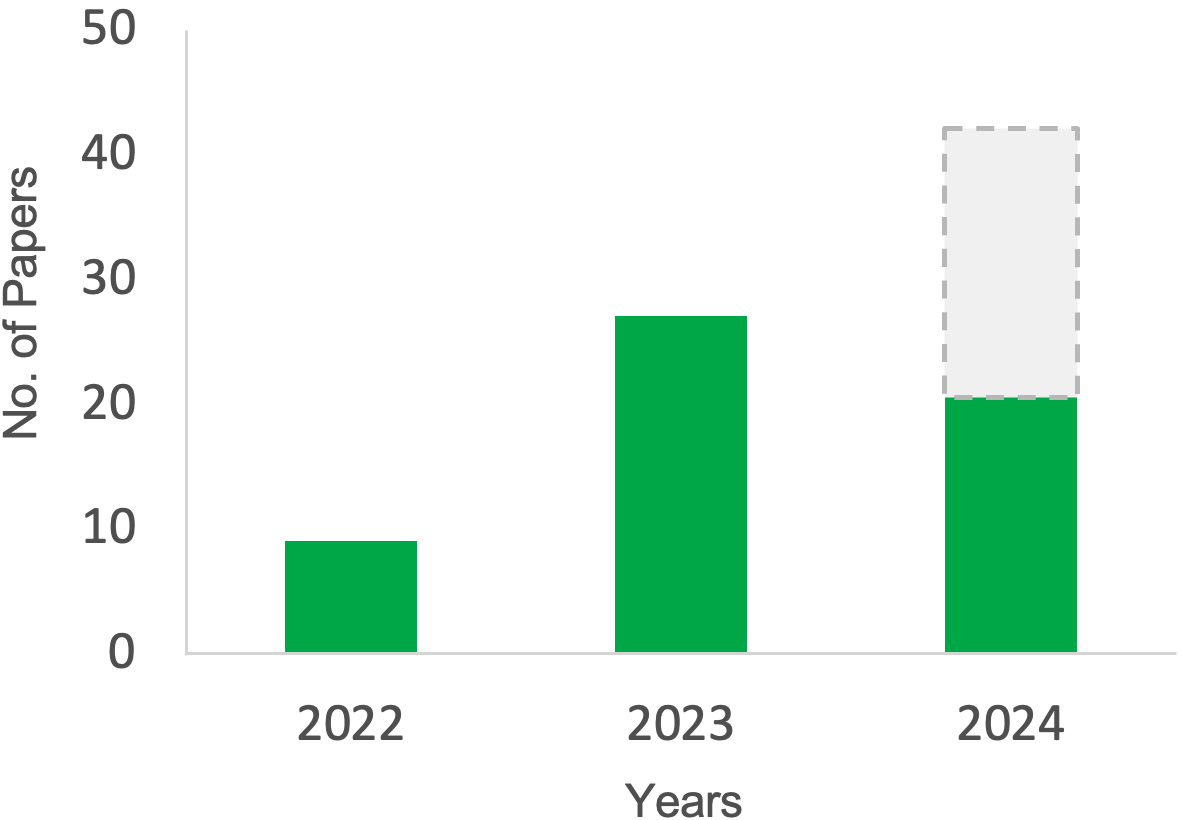}
    \caption{Shows yearly numbers of both published and preprint papers on diffusion models and representation learning. For 2024, the green bar indicates the number of papers collected up to and including June 2024, and the dashed grey bar indicates the projected number for the whole year.}
    \label{fig:trend_hot}
\end{figure}

Current state-of-the-art self-supervised representation learning approaches \cite{chen_simple_2020,grill_bootstrap_2020, caron_emerging_2021,asano2019self} have demonstrated great scalability. It is thus likely that diffusion models exhibit similar scaling properties~\cite{tian2024visual_var}. Controlled generation approaches like Classifier Guidance \cite{dhariwal_diffusion_2021} and Classifier-free Guidance \cite{ho2021classifier} used to obtain state-of-the-art generation results rely on annotated data, which represents a bottleneck for scaling up diffusion models. Guidance approaches that leverage representation learning and that are thus annotation-free offer a solution, potentially enabling diffusion models to train on much larger, annotation-free datasets.
\par
This survey paper aims to elucidate the relationship and interplay between diffusion models and representation learning. We highlight two central perspectives: Using diffusion models themselves for representation learning and using representation learning for improving diffusion models. We introduce a taxonomy of current approaches and derive generalized frameworks that demonstrate commonalities among current approaches.
\par
Interest in exploring the representation learning capabilities of diffusion models has been growing since the original formulation of diffusion models by \citet{sohl2015deep,ho2020denoising,song2021scorebased_sde}. As demonstrated in Fig.~\ref{fig:trend_hot}, we expect this trend to continue this year. The increased volume of published works on diffusion models and representation learning makes it more difficult for researchers to identify state-of-the-art approaches and stay on top of current developments. This can hinder progress in the space, which is why we feel a comprehensive overview and categorization is required.
\par
\begin{figure*}[t]
    \centering
    \includegraphics[width=\textwidth]{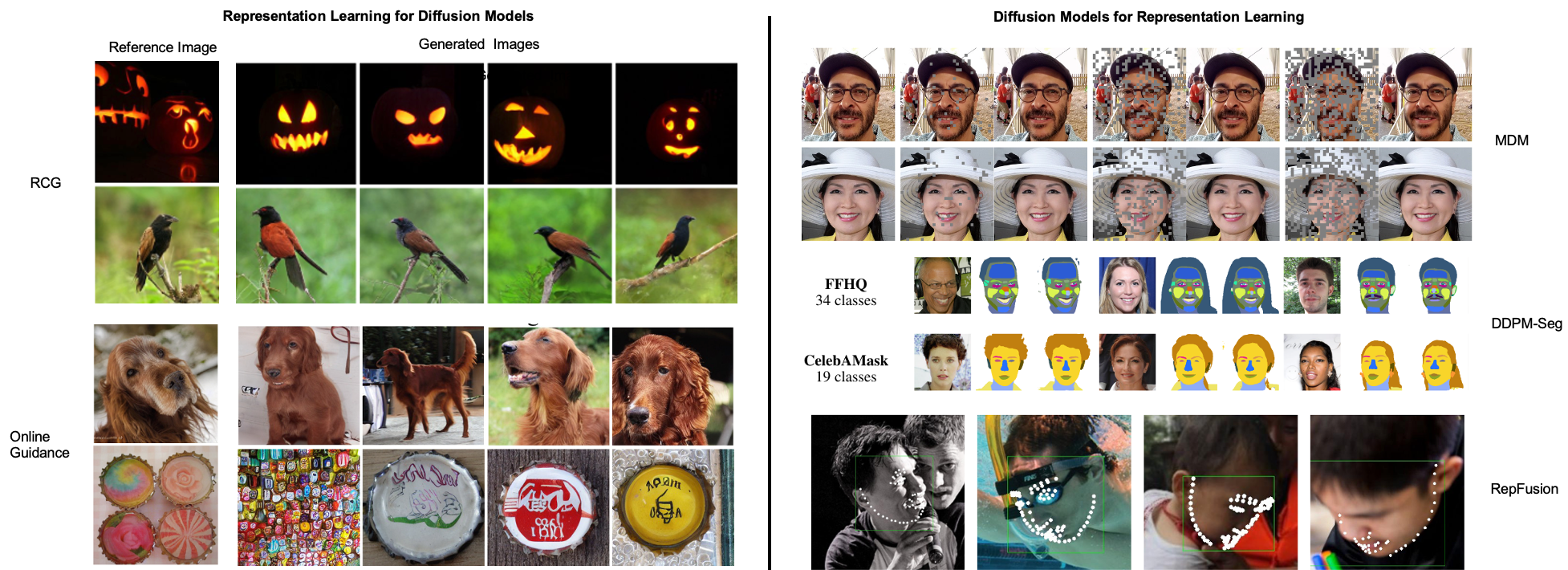}
    \caption{Left: Shows qualitative generation results from diffusion models conditioned using self-supervised guidance signals. Right: Shows qualitative results of downstream image tasks that leverage representations learned in training diffusion models. Adapted from \citet{li_return_2024}, \citet{hu_guided_2023},
    \citet{pan_masked_2024},
    \citet{baranchuk_label-efficient_2022},
    \citet{yang_diffusion_2023}.
    }
    \label{fig:qual_results}
\end{figure*}

Research on representation learning and diffusion models is in its infancy. Many of the current approaches rely on using diffusion models solely trained for generative synthesis for representation learning. We therefore hypothesize that there are significant opportunities for further progress in this area in the future and that diffusion models can increasingly challenge the current state-of-the-art in representation learning. Fig.~\ref{fig:qual_results} shows qualitative results from existing methods. We hope that this survey can contribute to advances in diffusion-based representation learning, by clarifying commonalities and differences among current approaches. In summary, the main contributions of this paper are the following:

\begin{itemize}
    \item \textbf{Comprehensive Overview}: Offers a thorough survey of the interplay between diffusion models and representation learning, providing clarity on how diffusion models can be used for representation learning and vice versa.
    \item \textbf{Taxonomy of Approaches}: We introduce a taxonomy of current approaches in diffusion-based representation learning, categorizing and highlighting commonalities and differences among them.
    \item \textbf{Generalized Frameworks}: The paper derives generalized frameworks for both diffusion model feature extraction and assignment-based guidance, offering a structured view on a large number of works on diffusion models and representation learning.
    \item \textbf{Future Directions}: We identify key opportunities for further progress in the field, encouraging the exploration of diffusion models and flow matching as a new state-of-the-art in representation learning.
\end{itemize}

\section{Background}\label{sec:background}

The following section outlines the required mathematical foundations of diffusion models. We also highlight current architecture backbones of diffusion models and provide a brief overview of sampling methods and conditional generation approaches.

\subsection{Mathematical Foundations}

Consider a set of training examples drawn from an underlying probability distribution $p(\mathbf{x})$. The idea behind generative diffusion models is to learn a denoising process that maps samples of random noise to novel images sampled from $p(\mathbf{x})$ \cite{po_state_2023}. To achieve this, images are corrupted by gradually adding different levels of Gaussian noise. Given an uncorrupted training sample $\mathbf{x}_0 \sim p(\mathbf{x})$, where index $0$ denotes the fact that the sample is not corrupted, the corrupted samples $\mathbf{x}_1, \mathbf{x}_2 ..., \mathbf{x}_T$ are generated according to a  Markovian process. One common choice for the transition kernel $p(\mathbf{x}_t|\mathbf{x}_{t-1})$ is the following:

\begin{equation}
\label{eq: eq1}
\begin{aligned}
    p(\mathbf{x}_t|\mathbf{x}_{t-1}) = \mathcal{N}\Big( & \mathbf{x}_t; \sqrt{1 - \beta_t} \mathbf{x}_{t-1}, &\beta_t \mathbf{I} \Big), \quad \\
    & \forall t \in \{1, \ldots, T\},
\end{aligned}
\end{equation}

where $T$ denotes the number of diffusion timesteps, $\beta_t$ is a time-dependent variance schedule and $\mathbf{I}$ is an identity matrix with dimensionality equal to $\mathbf{x}_0$ \cite{croitoru_diffusion_2023}. Note that other parametrizations of the transition kernel $p(\mathbf{x}_t|\mathbf{x}_{t-1})$ are also applicable in the same manner~\cite{kingma_variational_2021,zheng2023improved}. We proceed with the parametrization used in DDPMs \cite{ho2020denoising} to simplify the discussion moving forward. A noisy image $\mathbf{x}_t$ can be sampled directly from $\textbf{x}_0$ with the help of a reparametrization trick \cite{sohl2015deep} as follows:

\begin{equation}
    p(\mathbf{x}_t | \mathbf{x}_0) = \mathcal{N}\Big(\mathbf{x}_t; \sqrt{\Bar{\alpha}_t} \mathbf{x}_0;(1 - \Bar{\alpha}_t) \mathbf{I}\Big),
\end{equation}

where $\alpha_t := 1 - \beta_t$ and $\Bar{\alpha}_t := \prod^t_{i=1} \alpha_i$. Given the original input image $\mathbf{x}_0$, we can now obtain $\mathbf{x}_t$ in one step by sampling Gaussian vector $\boldsymbol{\epsilon}_t \sim \mathcal{N}(0, \mathbf{I})$ and applying:

\begin{equation}
    \mathbf{x}_t = \sqrt{\Bar{\alpha}_t} \mathbf{x}_0 + \sqrt{(1 - \Bar{\alpha}_t} )\boldsymbol{\epsilon}_t.
\end{equation}

We can generate novel samples from $p(\mathbf{x}_0)$ starting from a pure noise image $\mathbf{x}_T \sim \pi(\mathbf{x}_T) = \mathcal{N}(0, \mathbf{I})$ with dimensionality equivalent to the data and sequentially denoise it such that at every step, $p_\theta(\mathbf{x}_{t-1} | \mathbf{x}_t) = \mathcal{N}(\mathbf{x}_{t-1};\mu_\theta(\mathbf{x}_t, t), \Sigma_\theta(\mathbf{x}_t, t))$. In practice, this requires training a neural network $p_{\theta}(\mathbf{x}_{t-1} | \mathbf{x}_t)$ that predicts the mean $\mu_0(\mathbf{x}_t, t)$ and the covariance $\Sigma_{\theta}(\mathbf{x}_t, t)$ given a diffusion timestep $t$ and the noisy input image $\mathbf{x}_t$ \cite{yang_diffusion_2023-1}. Training this neural network with a maximum likelihood objective is intractable \cite{croitoru_diffusion_2023}, so the objective is amended to minimize a Variational Lower-Bound of the Negative Log-Likelihood instead \cite{sohl2015deep, ho2020denoising}:

\begin{equation}
\begin{aligned}
    \mathcal{L}_{vlb} = &- \log p_{\theta}(\mathbf{x}_0|\mathbf{x}_1) + D_{KL} \left( p(\mathbf{x}_T|\mathbf{x}_0) \| \pi(\mathbf{x}_T) \right) \\
    &+ \sum_{t > 1} D_{KL}\left( p(\mathbf{x}_{t-1}|\mathbf{x}_t, \mathbf{x}_0) \| p_{\theta}(\mathbf{x}_{t-1}|\mathbf{x}_t) \right),
\end{aligned}
\end{equation}

where $D_{KL}$ is the Kullback-Leibler divergence. This objective ensures that the neural network is trained to minimize the distance between $p_{\theta}(\mathbf{x}_{t-1} | \mathbf{x}_t)$ and the true posterior of the forward process when conditioned on $\mathbf{x}_0$. The denoising network is generally applied to parametrize the reverse mean $\mu_{\theta}(\mathbf{x}, t)$ of the distribution of the reverse transition $p_{\theta}(\mathbf{x}_{t-1} | \mathbf{x}_t):=\mathcal{N}(\mathbf{x}_{t-1}; \mu_{\theta}(\mathbf{x}_t, t), \Sigma_{\theta}(\mathbf{x}_t, t))$ \cite{chang_design_2023}. The true value of the reverse mean is a function of $\mathbf{x}_0$, which is unknown in the reverse process and must therefore be estimated using input timestep $t$ and the noisy data $\mathbf{x}_t$. Specifically, the reverse mean is formulated as the following:
\begin{equation}
\label{eq:rev_mean}
    \mu(\mathbf{x}_t, t) := \frac{\sqrt{\Bar{\alpha}_{t-1}} (1 - \Bar{\alpha}_{t-1})\mathbf{x}_t + \sqrt{\Bar{\alpha}_{t-1}} (1-\alpha_t)\mathbf{x}_0}{1 - \Bar{\alpha}_t},
\end{equation}

where the original data $\mathbf{x}_0$ is unavailable in the reverse process and must therefore be estimated. We denote the denoising network's prediction of the original data as $\hat{\mathbf{x}}_0$. This prediction $\hat{\mathbf{x}}_0$ can then be used to obtain $\mu_{\theta}(\mathbf{x}_t, t)$ using Equation~\ref{eq:rev_mean}. Parametrizing with $\hat{\mathbf{x}}_0$ directly is beneficial at the beginning of sampling, since predicting $\hat{\mathbf{x}}_0$ directly helps the denoising network to learn higher-level structural features \citep{luo_understanding_2022}. 
\par
\cite{ho2020denoising} suggest fixing the covariance $\Sigma_{\theta}(\mathbf{x}_t, t)$ to a constant value, which enables rewriting the parametrized reverse mean as a function of the added noise $\boldsymbol{\epsilon}(\mathbf{x}_t, t)$ instead of $\mathbf{x}_0$:
\begin{equation}
\label{eq: rev_mean_noise}
    \mu_{\theta}(\mathbf{x}_t, t) = \frac{1}{\sqrt{\alpha_t}} \left( \mathbf{x}_t - \frac{1 - \alpha_t}{\sqrt{1 - \Bar{\alpha}_t}} \boldsymbol{\epsilon}_\theta(\mathbf{x}_t, t). \right)
\end{equation}

This reparametrization allows for the derivation of a simplification of the objective $\mathcal{L}_{vlb}$ which we denote $\mathcal{L}_{simple}$ that measures the distance between the predicted noise $\boldsymbol{\epsilon}_\theta(\mathbf{x}_t, t)$ and the actual noise $\boldsymbol{\epsilon}_t$ as follows:
\begin{equation}
    \mathcal{L}_{simple} = \mathbb{E}_{t \sim [1, T]} \mathbb{E}_{\mathbf{x}_0 \sim p(\mathbf{x}_0)} \mathbb{E}_{\boldsymbol{\epsilon}_t \sim \mathcal{N}(0, \mathbf{I})} \left\| \boldsymbol{\epsilon}_t - \boldsymbol{\epsilon}_\theta(\mathbf{x}_t, t) \right\|^2.
\end{equation}

Instead of predicting the mean and covariance directly, the network is now parametrized to predict the added noise for a diffusion timestep and noisy image input. The reverse mean is obtained using Equation ~\ref{eq: rev_mean_noise}, and the covariance is fixed. Noise prediction networks have the benefit of being able to recover $\mathbf{x}_{t - 1}$ from $\mathbf{x}_{t}$ in the final sampling stages by predicting zero noise \citep{huang_variational_2021}. This is more difficult for direct parametrizations of $\hat{\mathbf{x}}_0$. There is therefore a tradeoff between the two, where direct parametrizations can be more beneficial for very noisy inputs in the initial sampling stages, and noise prediction parametrization can be beneficial in the latter sampling stages \cite{chang_design_2023}. 
\par
In efforts to improve sampling efficiency, \citet{salimans_progressive_2022} introduce velocity prediction as a further alternative parametrization. Velocity is a linear combination of the denoised input and the added noise, commonly defined as:
\begin{equation}
    \mathbf{v} = \Bar{\alpha}_t\epsilon - (1 - \Bar{\alpha}_t)\mathbf{x}_t.
\end{equation}

This parametrization combines benefits of both data and noise parametrizations, allowing the denoising network to flexibly learn noise prediction as well as reconstruction dynamics based on the signal-to-noise ratio. This parametrization has led to stable results in diffusion distillation approaches \citep{salimans_progressive_2022}, and can speed up generation \cite{benny_dynamic_2022}. 
\par
Recently, several works \cite{song_generative_2019,song2021scorebased_sde, po_state_2023,chen_wavegrad_2021} further propose to think of the noise in terms of continuous instead of discrete timesteps. Here, the diffusion process is expressed as a continuous time-dependent function $\sigma(t)$. Noise is gradually added whenever a sample $\mathbf{x}$ moves forward in time, and gradually removed if the image follows the reverse trajectory. More specifically, the diffusion process can be expressed using an Itô Stochastic Differential Equation (SDE) \cite{ito_stochastic_1950}, where the vector-valued drift coefficient $\mathbf{f}(\cdot, t): \mathbb{R}^d \to \mathbb{R}^d$ and the scalar-valued diffusion coefficient $g(\cdot): \mathbb{R} \to \mathbb{R}$ need to be selected when implementing a diffusion model:
\begin{equation}
    d \mathbf{x} = \mathbf{f}(\mathbf{x},t)dt + g(t)d \mathbf{w},
\end{equation}

where $\mathbf{w}$ is the standard Wiener process. There are two widely used choices of the SDE formulation used to model the diffusion process. The first is the Variance-Preserving (VP) SDE, used in the work of \citet{ho2020denoising} which is given by $\mathbf{f}(\mathbf{x},t)= -\frac{1}{2}\beta(t)\mathbf{x}$ and $g(t) = \sqrt{\beta(t)}$, where $\beta(t) = \beta_t$ as $T$ goes to infinity. Note that this is equivalent to the continuous formulation of the DDPM parametrization in Equation~\ref{eq: eq1}. The second is the Variance-Exploding (VE) SDE \cite{song_generative_2019}, resulting from a choice of $\mathbf{f}(\mathbf{x},t)=0$ and $g(t)=\sqrt{2 \sigma(t) \frac{d\sigma(t)}{dt}}$. The VE SDE gets its name since the variance continually increases with increasing $t$, whereas the variance in the VP SDE is bounded \cite{song2021scorebased_sde}. Anderson \cite{anderson_reverse-time_1982} derives an SDE that reverses a diffusion process, which results in the following when applied to the Variance Exploding SDE:
\begin{equation}
    d\mathbf{x} = -2\sigma(t) \frac{d\sigma(t)}{dt} \nabla_\mathbf{x} \log p(\mathbf{x}; \sigma(t)) \, dt + \sqrt{2\sigma(t) \frac{d\sigma(t)}{dt}} \, d\mathbf{w}.
\end{equation}

$\nabla_\mathbf{x} \log p(\mathbf{x}; \sigma(t))$ is known as the score function. This score function is generally not known, so it needs to be approximated using a neural network. A neural network $D(\mathbf{x}; \sigma)$ that minimizes the L2-denoising error can be used to extract the score function since $\nabla_{\mathbf{x}} \log p(\mathbf{x}; \sigma(t)) = \frac{D(\mathbf{x}; \sigma) - \mathbf{x}}{\sigma^2}$.  This idea is known as Denoising Score Matching \citep{vincent2011connection}. 


\subsection{Backbone Architectures}

\begin{figure*}
    
    \centering
    \includegraphics[scale=0.45]{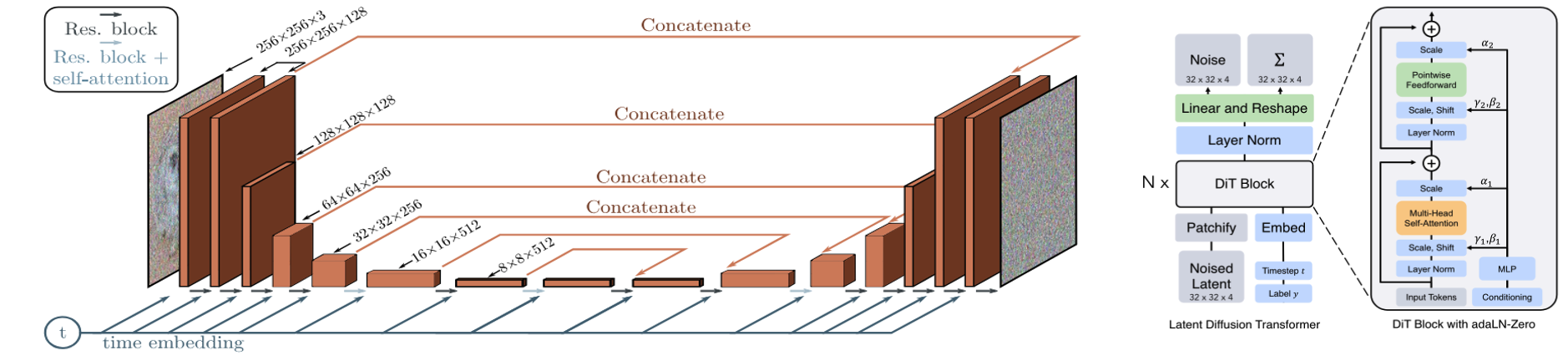}
    \caption{Left: An exemplary visualization of the U-Net architecture~\cite{ronneberger_u-net_2015}. Consists of an encoder and a decoder, with residual connections that preserve gradient flow and low-level input details. Adapted from \cite{prince_understanding_2023}. Right: An exemplary visualization of the DiT architecture. Shows the high-level architecture, as well as a breakdown of the adaLN-Zero DiT block. Adapted from \citet{peebles_scalable_2023}.
    }
    \label{fig:backbones}
\end{figure*}

We outline the mathematical foundations of diffusion models in Section 2.1. Since denoising prediction networks are generally parametrized by parameters $\theta$, we discuss the formulation of $\theta$ by several neural network architectures in the following section. All of these network architectures map from the same input space to the same output space.
\par
\citet{ho2020denoising} use a U-Net backbone similar to an unmasked PixelCNN++ \citep{salimans_pixelcnn_2017} to approximate the score function. This U-Net architecture, originally used in semantic segmentation approaches \cite{ronneberger_u-net_2015,long_fully_2015, chen_semantic_2016, chen_deeplab_2017}, is based on a Wide ResNet \citep{zagoruyko_wide_2016} and takes a noisy image and the diffusion timestep $t$ as input, encodes the image to a lower-dimensional representation, and outputs the noise prediction for that image and noise level. The U-Net consists of an encoder and a decoder with residual connections between blocks that preserve gradient flow and help recover fine-grained details lost in the compressed representation. The encoder consists of a series of residual and self-attention blocks and downsamples the input image to a low-dimensional representation. The decoder mirrors this structure, gradually upsampling the low-dimensional representation to match the input dimensionality. The diffusion timestep $t$ is specified by adding a sinusoidal positional embedding in each residual block \citep{ho2020denoising} that scales and shifts the input features, enhancing the network's ability to capture temporal dependencies.
\par
DDPMs operate in the pixel space, making their training and inference computationally expensive. \citet{rombach2022high_latentdiffusion_ldm} address this by proposing Latent Diffusion Models (LDMs), which operate in the latent space of a pre-trained variational autoencoder. The diffusion process is applied to the generated representation as opposed to the image directly, leading to computational benefits without sacrificing generation quality. While the authors introduce additional cross-attention mechanisms to allow for more flexible conditioned generation, the denoising network backbone remains very close to the DDPM U-Net architecture. 
\par
Recent advances in the use of transformer architectures for vision tasks like ViT \citep{dosovitskiy2020image_vit} have led to the adoption of transformer-based architectures for diffusion models. \citet{peebles_scalable_2023} propose Diffusion Transformers (DiT), a diffusion model backbone architecture that is largely inspired by ViTs, and demonstrates state-of-the-art generation performance on ImageNet when combined with the LDM framework. Following ViT, DiTs work by transforming input images into a sequence of patches, which are converted into a sequence of tokens using a "patchify" layer. After adding ViT-style positional embeddings to all input tokens, the tokens are fed through a series of transformer blocks. These blocks are equivalent to standard ViT blocks that take additional conditional information such as the diffusion timestep $t$ and a conditioning signal $\mathbf{c}$ as inputs. A detailed overview of their structure can be seen in Fig~\ref{fig:backbones}.
\par
U-ViTs \cite{bao_all_2023} combine the U-Net and ViT backbones into a unified backbone. U-ViTs follow the design methodology of transformers in tokenizing time, conditioning and image inputs, but additionally employ long skip connections between shallow and deep layers. These skip connections provide shortcuts for low-level features and therefore stabilize training of the denoising network \cite{bao_all_2023}. Works utilizing U-ViT-based backbones \cite{hoogeboom_simple_2023,bao_one_2023} achieve results on par with U-Net CNN-based architectures, demonstrating their potential as a viable alternative to other denoising network backbones. 

\subsection{Diffusion Model Guidance}
\begin{table}
\caption{An overview of different diffusion model guidance approaches. Self-guidance \cite{hu_self-guided_2023} and \cite{hu_guided_2023} are both classifier and annotation-free, and online guidance facilitates online learning.}
\label{tab:guidance_overview}
\centering
\resizebox{0.49\textwidth}{!}{
\begin{tabular}{l|c|c|c}
\toprule
\textbf{Approach} & \textbf{Classifier-Free} & \textbf{Annotation-Free} & \textbf{Online Learning} \\
\midrule
Classifier Guidance~\cite{dhariwal2021diffusion_beat} & \textcolor{red}{\ding{55}} & \textcolor{red}{\ding{55}} & \textcolor{red}{\ding{55}}  \\
Classifier-free Guidance~\cite{ho2021classifier} & \textcolor{green}{\ding{51}} & \textcolor{red}{\ding{55}} & \textcolor{red}{\ding{55}} \\
Self-guidance~\cite{hu_self-guided_2023,li_return_2024} & \textcolor{green}{\ding{51}} & \textcolor{green}{\ding{51}} & \textcolor{red}{\ding{55}} \\
Online guidance~\cite{hu_guided_2023} & \textcolor{green}{\ding{51}} & \textcolor{green}{\ding{51}} & \textcolor{green}{\ding{51}} \\
\bottomrule
\end{tabular}
} 
\end{table}

Recent improvements in image generation results have largely been driven by improved guidance approaches. The ability to control generation by passing user-defined conditions is an important property of generative models, and guidance describes the modulation of the strength of the conditioning signal within the model. Conditioning signals can have a wide range of modalities, ranging from class labels, to text embeddings to other images. A simple method to pass spatial conditioning signals to diffusion models is to simply concatenate the conditioning signal with the denoising targets and then pass the signal through the denoising network~\cite{hu_self-guided_2023, bao_all_2023}. Another effective approach uses cross-attention mechanisms, where a conditioning signal $\mathbf{c}$ is preprocessed by an encoder to an intermediate projection $E(\mathbf{c})$, and then injected into the intermediate layer of the denoising network using cross-attention \cite{saharia2022photorealistic,hu_zigma_2024}. These conditioning approaches alone do not leave the possibility to regulate the strength of the conditioning signal within the model. Diffusion model guidance has recently emerged as an approach to more precisely trade-off generation quality and diversity.
\par
\citet{dhariwal2021diffusion_beat} use classifier guidance, a compute-efficient method leveraging a pre-trained noise-robust classifier to improve sample quality. Classifier guidance is based on the observation that a pre-trained diffusion model can be conditioned using the gradients of a classifier parametrized by $\phi$ outputting $p_{\phi}(\mathbf{c} | \mathbf{x_t}, t)$. The gradients of the log-likelihood of this classifier $\nabla_{\mathbf{x_t}} \log p_{\phi} (\mathbf{c} | \mathbf{x_t}, t)$ can be used to guide the diffusion process towards generating an image belonging to class label $\mathbf{y}$. The score estimator for $p(x | \mathbf{c})$ can be written as
\begin{equation}
    \nabla_{\mathbf{x_t}} \log \left( p_{\theta}(\mathbf{x_t}) p_{\phi}(\mathbf{c} | \mathbf{x_t}) \right) = \nabla_{\mathbf{x_t}} \log p_{\theta}(\mathbf{x_t}) + \nabla_{\mathbf{x_t}} \log p_{\phi}(\mathbf{c} | \mathbf{x_t}).
\end{equation}

By using Bayes' theorem, the noise prediction network can then be rewritten to estimate:
\begin{equation}
    \hat{\epsilon}_{\theta}(\mathbf{x_t}, \mathbf{c}) = \epsilon_{\theta}(\mathbf{x_t}, \mathbf{c}) - w \sigma_t \nabla_{\mathbf{x_t}} \log p_{\phi}(\mathbf{c} | \mathbf{x_t}),
\end{equation}

where the parameter $w$ modulates the strength of the conditioning signal. Classifier guidance is a versatile approach that increases sample quality, but it is heavily reliant on the availability of a noise-robust pre-trained classifier, which in turn relies on the availability of annotated data, which is not available in many applications.
\par
To address this limitation, Classifier-free guidance (CFG) \cite{ho2021classifier} eliminates the need for a pre-trained classifier. CFG works by training an unconditional diffusion model parametrized by $\epsilon_{\theta}(\mathbf{x_t}, t, \phi)$ together with a conditional model parametrized by $\epsilon_{\theta}(\mathbf{x_t}, t, \mathbf{c})$. For the unconditional model, a null input token $\phi$ is used as a conditioning signal $\mathbf{c}$. The network is trained by randomly dropping out the conditioning signal with probability $p_{\text{uncond}}$. Sampling is then performed using a weighted combination of conditional and unconditional score estimates:
\begin{equation}
    \Tilde{\epsilon}_{\theta}(\mathbf{x_t}, \mathbf{c}) = (1 + w)\epsilon_{\theta}(\mathbf{x_t}, \mathbf{c}) - w\epsilon_{\theta}(\mathbf{x_t}, \phi).
\end{equation}

This sampling method does not rely on the gradients of a pre-trained classifier but still requires an annotated dataset to train the conditional denoising network. Fully unconditional approaches have yet to match classifier-free guidance, though recent works using diffusion model representations for self-supervised guidance show promise \cite{li_return_2024,hu_guided_2023}. These methods do not need annotated data, allowing the use of larger unlabelled datasets.
\par
Table~\ref{tab:guidance_overview} shows the requirements of current guidance methods. While classifier and classifier-free guidance improve generation results, they require annotated training data. Self-guidance and online guidance are fully self-supervised alternatives that achieve competitive performance without annotations.
\par
Classifier and classifier-free guidance are controlled generation methods that rely on conditional training. Training-free approaches modify the generation process of a pre-trained model by binding multiple diffusion processes \cite{bar-tal_multidiffusion_2023} or using time-independent energy functions \cite{yu_freedom_2023}. Other controlled generation methods take a variational perspective \cite{graikos_diffusion_2022,mardani_variational_2024,wallace_end--end_2023,samuel_generating_2024}, treating controlled generation as a source point optimization problem \cite{ben-hamu_d-flow_2024}. The goal is to find samples \( \mathbf{x} \) that minimize a loss function \( \mathcal{L}(\mathbf{x}) \) and are likely under the model's distribution \( p \). The optimization is formulated as \( \min_{\mathbf{x_0}}\mathcal{L}(\mathbf{x}) \), where \( \mathbf{x_0} \) is the source noise point. The loss function \( \mathcal{L}(\mathbf{x}) \) can be modified for conditional sampling to generate a sample belonging to a particular class \( \mathbf{y} \).

\section{Methods}\label{sec:methods}

\begin{table*}
\centering
\caption{Summary of the methods using diffusion models for representation learning.} 
\label{tab:scanning} 
\begin{tabular}{l|l|r}
\toprule
\textbf{Paradigm} & \textbf{Downstream Task} & \textbf{Method} \\
\midrule
\multirow{4}{*}{Generative Augmentation} 
& \multirow{2}{*}{Classification} & Generative Augmentation~\cite{ayromlou_can_2024} \\
& & MA-ZSC~\cite{shipard2023diversity} \\
\cmidrule{2-3}
& \multirow{1}{*}{Semantic Segmentation} & ScribbleGen~\cite{schnell_scribblegen_2024} \\
\midrule
\multirow{17}{*}{Leveraging Intermediate Activations} 
& \multirow{3}{*}{Classification} & GDC~\cite{mukhopadhyay_diffusion_2023} \\
& & DifFormer~\cite{mukhopadhyay_text-free_2023} \\ 
& & DDAE~\cite{xiang_denoising_2023} \\
\cmidrule{2-3}
& \multirow{2}{*}{Semantic Segmentation} & DDPM-Seg~\cite{baranchuk_label-efficient_2022} \\
& & VDM~\cite{zhao_unleashing_2023} \\
\cmidrule{2-3}
& Panoptic Segmentation & ODISE~\cite{xu_open-vocabulary_2023} \\
\cmidrule{2-3}
& \multirow{5}{*}{Semantic Correspondence} & DIFT~\cite{tang_emergent_2023} \\
& & SD+DINO~\cite{zhang_tale_2023} \\
& & Diffusion Hyperfeatures~\cite{luo_diffusion_2024} \\
& & SD4Match~\cite{li_sd4match_2024} \\
& & USCSD~\cite{hedlin_unsupervised_2023} \\
\cmidrule{2-3}
& \multirow{1}{*}{Depth Estimation} & VDM~\cite{zhao_unleashing_2023} \\
\cmidrule{2-3}
& \multirow{2}{*}{Image Editing} &
P2PCAC~\cite{hertz_prompt--prompt_2022} \\
& & Plug-and-Play Diffusion Features~\cite{tumanyan_plug-and-play_2022} \\
\midrule
\multirow{10}{*}{Diffusion Model Reconstruction} 
& \multirow{3}{*}{Classification} & SODA~\cite{hudson_soda_2024} \\
& & l-DAE~\cite{chen_deconstructing_2024} \\
& & DiffMAE~\cite{wei_diffusion_2023} \\
\cmidrule{2-3}
&\multirow{1}{*}{Semantic Segmentation} & MDM~\cite{pan_masked_2024} \\
\cmidrule{2-3}
& \multirow{2}{*}{Image Editing} & DiffAE~\cite{preechakul2022diffusion_autoencoder} \\
& & PDAE~\cite{zhang_unsupervised_2022} \\
\cmidrule{2-3}
& \multirow{2}{*}{Image Interpolation} & InfoDiffusion~\cite{wang_infodiffusion_2023} \\
& & SmoothDiffusion~\cite{guo_smooth_2024} \\
\midrule
\multirow{3}{*}{Diffusion Model Knowledge Transfer} 
& \multirow{3}{*}{Classification} & DiffusionClassifier~\cite{li_your_2023} \\
& & RepFusion~\cite{yang_diffusion_2023} \\
& & DreamTeacher~\cite{li_dreamteacher_2023} \\
\midrule
\multirow{3}{*}{Joint Diffusion Models} 
& \multirow{2}{*}{Classification} & JDM~\cite{deja_learning_2023} \\
& & HybViT~\cite{yang_your_2022} \\
\cmidrule{2-3}
& \multirow{1}{*}{Semantic Segmentation}
& ADDP~\cite{tian_addp_2024} \\
\bottomrule
\end{tabular}
\end{table*}

Having covered the main preliminaries for diffusion models, we outline a series of methods related to diffusion models and representation learning in the following section. In subsection \ref{subsec:d4r} we describe and categorize current frameworks utilizing representations learned by pre-trained diffusion models for downstream recognition tasks. In subsection \ref{subsec:r4d}, we describe methods that leverage advances in representation learning to improve diffusion models themselves.

\subsection{Diffusion Models for Representation Learning}
\label{subsec:d4r}

Learning useful representations is one of the main motivations for designing architectures like VAEs \cite{kingma2013auto_vae, kingma_introduction_2019} and GANs \cite{karras_style-based_2019,brock2018large_biggan}. Contrastive learning approaches, where the goal is to learn a feature space in which representations of similar images are very close together, and vice versa for dissimilar images (e.g. SimCLR~\cite{chen2020big_simclr_v2}, MoCo \citep{he2020momentum_moco}), have also led to significant advances in representation learning. These contrastive methods are not fully self-supervised however, since they require supervision in the form of augmentations that preserve the original content of the image. 
\par
Diffusion models offer a promising alternative to these approaches. While diffusion models are primarily designed for generation tasks, the denoising process encourages the learning of semantic image representations \cite{baranchuk_label-efficient_2022}, that can be used for downstream recognition tasks. The diffusion model learning process is similar to the learning process of Denoising Autoencoders (DAE) \cite{vincent_extracting_2008,bengio_generalized_2013}, which are trained to reconstruct images corrupted by adding noise. The main difference is that diffusion models additionally take the diffusion timestep $t$ as input, and can thus be viewed as multi-level DAEs with different noise scales \cite{xiang_denoising_2023}. Since DAEs learn meaningful representations in the compressed latent space, it is intuitive that diffusion models exhibit similar representation learning capabilities. We outline and discuss current approaches in the following section.

\subsubsection{Leveraging intermediate activations} 

\citet{baranchuk_label-efficient_2022} investigate the intermediate activations from the U-Net network that approximates the Markov step of the reverse diffusion process in DDPMs \cite{dhariwal2021diffusion_beat}. They show that for certain diffusion timesteps, these intermediate activations capture semantic information that can be used for downstream semantic segmentation. The authors take a noise-predictor network $\epsilon_{\theta}(\mathbf{x}_t, t)$ trained on the LSUN-Horse \cite{yu15lsun} and FFHQ-256 \cite{karras_style-based_2019} datasets and extract feature maps produced by one of the network's 18 decoder blocks for label-efficient downstream segmentation tasks. Selecting the ideal diffusion timestep and decoder block activation to extract is non-trivial. To understand the efficacy of pixel-level representations of different decoder blocks, the authors train a multi-layer perceptron (MLP) to predict the semantic label from features produced by different decoder blocks on a specific diffusion step $t$. The representations from a fixed set of blocks $B$ of the pre-trained U-Net decoder and higher diffusion timesteps are upsampled to the image size using bilinear interpolation and concatenated. The obtained feature vectors are then used to train an ensemble of independent MLPs which predict a semantic label for each pixel. The final prediction is obtained by majority voting. This method, denoted DDPM-Seg, outperforms baselines that exploit alternative generative models and achieves segmentation results competitive with MAE \cite{he2022masked_mae}, illustrating that intermediate denoising network activations contain semantic image features.
\par
\citet{xiang_denoising_2023} extend this approach to further architectures and image recognition on CIFAR-10 and Tiny-ImageNet. They investigate the discriminative efficacy of extracted features for different backbones (U-Net and DiT \cite{peebles_scalable_2023}) under different frameworks (DDPM and EDM \cite{karras2022elucidating}). The relationship between feature quality and layer-noise combinations is evaluated through grid search, where the quality of feature representations is determined using linear probing. The best-performing features lie in the middle of up-sampling using relatively small noising levels, which is in line with conclusions drawn in DDPM-Seg \cite{baranchuk_label-efficient_2022}. Benchmark comparisons against diffusion-based methods like HybViT \cite{yang_your_2022} and SBGC \cite{zimmermann_score-based_2021} on CIFAR-10 and Tiny-ImageNet \cite{deng2009imagenet} show that EDM-based Denoising Diffusion Autoencoders (DDAEs) outperform previous supervised and unsupervised diffusion-based methods on both generation and recognition, especially after fine-tuning. Benchmarking against contrastive learning methods shows that the EDM-based DDAE is comparable with Sim-CLRs considering model sizes, and outperforms SimCLRs with comparable parameters on CIFAR-10 and Tiny-ImageNet. 
\par
ODISE \cite{xu_open-vocabulary_2023} is a related approach that unites text-to-image diffusion models with discriminative models to perform panoptic segmentation \cite{kirillov_panoptic_2019, kirillov_panoptic_2019-1}, a segmentation approach unifying instance and semantic segmentation into a common framework for comprehensive scene understanding. ODISE extracts the internal features of a pre-trained text-to-image diffusion model. These features are input to a mask generator trained on annotated masks. A mask classification module then categorizes each generated binary mask into an open vocabulary category by relating the predicted mask's diffusion features with text embeddings of object category names. The authors use the Stable Diffusion U-Net DDPM backbone and extract features by computing a single forward pass and extracting the intermediate activations $f = \text{UNet}(\mathbf{x}_t, \tau(s), t)$ where $\tau(s)$ is an encoded representation of the image caption $s$ obtained leveraging a pre-trained text encoder $\tau$. Interestingly, the authors obtain the best results using $t = 0$, whereas previous methods obtain better results using higher diffusion timesteps. To overcome reliance on available image captions, \citet{xu_open-vocabulary_2023} additionally train an MLP-based implicit captioner that computes an implicit text embedding from the image itself. ODISE establishes a new state-of-the-art in open-vocabulary segmentation and is a further example of the rich semantic representations learned by denoising diffusion models.
\par
\citet{mukhopadhyay_diffusion_2023} also propose leveraging intermediate activations from the unconditional ADM U-Net architecture \cite{dhariwal2021diffusion_beat} for ImageNet classification. The methodology for layer and timestep selection is similar to previous approaches. Additionally, the impact of different sizes for feature map pooling is evaluated and several different lightweight architectures for classification (including linear, MLP, CNN, and attention-based classification heads) are used. Feature quality is found to be mostly insensitive to pooling size, and is mostly dependent on time steps and the selected block number. Their approach, which we term guided diffusion classification (GDC), achieves competitive performance against other unified models, namely BigBiGAN \cite{donahue2019large_bigbigan} and MAGE \cite{li_mage_2023}. The attention-based classification heads perform best on ImageNet-50, but perform poorly on Fine-Grained Visual Classification datasets, indicating their reliance on a large amount of available data.
\par
In a continuation of their previous work, \citet{mukhopadhyay_text-free_2023} extend this approach by introducing two methods for more fine-grained block and denoising time step selection. The first is DifFormer \cite{mukhopadhyay_text-free_2023}, an attention mechanism replacing the fixed pooling and linear classification head from \cite{mukhopadhyay_diffusion_2023} with an attention-based feature fusion head. This fusion head is designed to replace the fixed flattening and pooling operation required to generate vector feature representations from the U-Net CNN used in the GDC approach with a learnable pooling mechanism. The second mechanism is DifFeed \cite{mukhopadhyay_text-free_2023}, a dynamic feedback mechanism that decouples the feature extraction process into two forward passes. In the first forward pass, only the selected decoder feature maps are stored. These are fed to an auxiliary feedback network that learns to map decoder features to a feature space suitable for adding them to the encoder blocks of corresponding blocks. In the second forward pass, the feedback features are added to the encoder features, and the DifFeed attention head is used on top of those second forward pass features. These additional improvements further increase the quality of learned representations and improve ImageNet and fine-grained visual classification performance. 
\par
The previously described diffusion representation learning methods focus on segmentation and classification, which are only a subset of downstream recognition tasks. Correspondence tasks are another subset that generally involves identifying and matching points or features between different images. The problem setting is as follows: Consider two images $\mathbf{I}_1$ and $\mathbf{I}_2$ and a pixel location $p_1$ in $\mathbf{I}_1$. A correspondence task involves finding the corresponding pixel location $p_2$ in $\mathbf{I}_2$. The relationship between $p_1$ and $p_2$ can be semantic (pixels that contain similar semantics), geometrical (pixels that contain different views of an object) or temporal (pixels that contain the same object deforming over time). DIFT (Diffusion Features) \cite{tang_emergent_2023} is an approach leveraging pre-trained diffusion model representations for correspondence tasks. DIFT also relies on extracting diffusion model features. Similarly to previous approaches, diffusion timestep and network layer numbers used for extraction are an important consideration. The authors observe more semantically meaningful features for large diffusion timesteps and earlier network layer combinations, whereas lower-level features are captured in smaller diffusion timesteps and later denoising network layers. DIFT is shown to outperform other self-supervised and weakly-supervised methods across a range of correspondence tasks, showing on-par performance with state-of-the-art methods on semantic correspondence specifically.
\par
\citet{zhang_tale_2023} evaluate how learned diffusion features relate across multiple images, instead of focusing on downstream tasks for single images. To investigate this, they employ Stable Diffusion features for semantic correspondence as well. The authors observe that Stable Diffusion features have a strong sense of spatial layout, but sometimes provide inaccurate semantic matches. DINOv2 \cite{oquab_dinov_2024}, a method for self-supervised representation learning using knowledge distillation and vision transformers, produces more sparse features that provide more accurate matches. \citet{zhang_tale_2023} therefore propose to combine the two features and employ zero-shot evaluation of nearest neighbor search on the combined features to achieve state-of-the-art performance on several semantic correspondence datasets like SPair-71k and TSS.
\par
SD4Match \cite{li_sd4match_2024} builds on this approach by using various prompt tuning and conditioning techniques. One method, SD4Match-Class, fine-tunes prompt embedding $\Theta$ for each semantic class using a semantic matching loss \cite{li_simsc_2023}. Given images $\mathbf{I}_t^{A}$ and $\mathbf{I}_t^{B}$, the Stable Diffusion U-Net $f(\cdot)$ extracts feature maps $\mathbf{F}_t^{A}$ and $\mathbf{F}_t^{B}$ by $\mathbf{F}_t = f(\mathbf{I}_t, t, \boldsymbol{\Theta})$. Correspondence points are predicted by normalizing feature maps and computing a correlation map, which is converted to a probability distribution using a softmax operation. Additionally, \citet{li_sd4match_2024} propose conditioning prompts on input images using a Conditional Prompting Module (CPM), which includes a DINOv2 feature extractor, linear layers, and an adaptive MaxPooling layer. The conditioning embedding $\boldsymbol{\Theta}_{\text{cond}}$ is formed by concatenating feature representations and projecting them to the prompt embedding dimension. The final prompt $\boldsymbol{\Theta}_{\text{AB}}$ is obtained by appending $\boldsymbol{\Theta}_{\text{cond}}$ to a global prompt $\boldsymbol{\Theta}_{\text{global}}$. This method sets new benchmark accuracies on SPair-71k \cite{min_spair-71k_2019}, PF-Willow, and PF-Pascal \cite{ham_proposal_2017}, surpassing methods like DIFT \cite{tang_emergent_2023} and SD+DINO \cite{zhang_tale_2023}
\par
\citet{luo_diffusion_2024} introduce Diffusion Hyperfeatures, a framework designed to consolidate multiple intermediate activation maps across diffusion timesteps for downstream recognition. Activations are consolidated using an interpretable aggregation network, that takes the collection of intermediate feature maps as input and produces a single feature descriptive feature map as output. While other approaches manually select fixed diffusion timesteps and activations from a pre-determined number of intermediate network layers, Diffusion Hyperparameters cache all feature maps across all layers and timesteps in the diffusion process to generate a dense set of activations. This high dimensional set of activations is upsampled, passed through a bottleneck layer $B$ and weighed with a unique learnable mixing weight $w_{l,s}$ for each layer and timestep combination. The final diffusion hyperfeatures take on the form 
\begin{equation}
    \sum_{s=0}^{S} \sum_{l=1}^{L} w_{l,s} B_{l}(\mathbf{r}_{l,s}),
\end{equation}

where $L$ is the number of layers, $S$ is a subsample of the number of diffusion timesteps and $r$ is an activation feature map. Bottleneck layers and mixing weights are finetuned on the specific downstream task. Similar to previous approaches, Diffusion Hyperfeatures is used for semantic correspondence. The authors extract activations from Stable-Diffusion and tune the aggregation network on a subset of SPair-71k. Diffusion Hyperfeatures outperforms models that use self-supervised descriptors or supervised hypercolumns on the SPair-71k and CUB datasets.
\par
\citet{hedlin_unsupervised_2023} focus on optimizing the prompt embeddings by exploiting intermediate attention maps specifically. Given a certain input text prompt, these attention activation maps correspond to the semantics of the prompt. Instead of optimizing a global or a class-dependent prompt embedding $\Theta$ using the semantic loss, \citet{hedlin_unsupervised_2023} optimize the embedding to maximize the cross-attention at the location of interest. Locating corresponding points in a second image then comes down to conditioning on the optimized prompt, and selecting the point with the pixel attaining the maximum attention map value within the target image. Note that this approach does not utilize supervised training specific to semantic correspondence. However, they require test-time optimization which is costly. Text prompts are optimized using an off-the-shelf diffusion model without fine-tuning. Several further works building on aforementioned approaches \cite{mariotti_improving_2024, zhang_telling_2024} exist, showing that exploiting pre-trained diffusion models for semantic correspondence remains a promising application of diffusion models.
\par
\citet{zhao_unleashing_2023} propose Visual Perception with a pre-trained Diffusion Model (VDM), a framework closely related to USCSD \cite{hedlin_unsupervised_2023} that employs a text feature refinement network as well as an additional recognition encoder for semantic segmentation and depth estimation. Here, the denoising network is fed with refined text representations as well as an input image, and the resulting feature maps as well as the cross-attention maps between the text and image features are used to provide guidance for a decoder. To achieve this, the prediction model is written as $p_{\phi}(\mathbf{y} | \mathbf{x}, \mathcal{S})$, where $\mathcal{S}$ represents the set of all category labels of the downstream task. The prediction model is implemented as the following:
\begin{equation}
    p_{\phi}(\mathbf{y}|\mathbf{x}, S) = p_{\phi_3}(\mathbf{y}|\mathcal{F}) p_{\phi_2}(\mathcal{F}|\mathbf{x}, \mathcal{C}) p_{\phi_1}(\mathcal{C}|\mathcal{S}),
\end{equation}

where $\mathcal{F}$ denotes the set of feature maps and $\mathcal{C}$ denotes the text features. Here, $p_{\phi_1}(\mathcal{C}|\mathcal{S})$ denotes a text adapter consisting of a two-layer MLP that refines the text features obtained by applying the CLIP text encoder to a text template of $\texttt{"a photo of a [CLS]"}$. $p_{\phi_2}(\mathcal{F}|\mathbf{x})$ extracts the feature maps from the denoising network given the input image $\mathbf{x}$ and the set of refined text features $\mathcal{C}$. The authors use $t=0$ when feeding the denoising network the latent representation of the input image generated by using the VQGAN encoder \cite{esser_taming_2021} to obtain feature maps $\mathcal{F}$. Finally, $p_{\phi_3}(\mathbf{y}|\mathcal{F})$ serves as a light-weight prediction head implemented as a semantic feature pyramid network \cite{kirillov_panoptic_2019} that is adapted to the downstream task. VDM is evaluated on semantic segmentation and depth estimation, and achieves highly competitive performance and fast convergence compared to methods with other pre-training paradigms.
\par
A more indirect application of text-to-image diffusion model representations is instructional image editing \cite{brooks_instructpix2pix_2023,geng_instructdiffusion_2024,li_moecontroller_2024}, where the desired image edit is described by a natural language instruction rather than a description of the desired new image \cite{huang_diffusion_2024}. Prompt-based image editing is challenging since small changes in the textual prompt can lead to vastly different generation outcomes. \cite{hertz_prompt--prompt_2022} propose a textual editing method for pre-trained text-conditioned diffusion models that leverages the semantic strength of the intermediate cross-attention layers in the denoising backbone. This approach is based on a key observation also employed in \cite{hedlin_unsupervised_2023,zhao_unleashing_2023}: Cross-attention maps contain rich information on the spatial layout and geometry of the generated image. Injecting the cross-attention layers obtained when generating an image $\mathcal{I}$ into the generation process of the edited image $\mathcal{I^*}$ ensures that the edited image preserves the original spatial layout. \citet{hertz_prompt--prompt_2022} use Imagen \cite{saharia_photorealistic_2022} to conduct experiments and demonstrate promising results on text-only localized editing, global editing, and real image editing. Following works like Plug-and-play Diffusion Features \cite{tumanyan_plug-and-play_2022} further improve upon this by leveraging all intermediate activation maps to enable instructional image editing. Other techniques like TokenFlow \cite{geyer_tokenflow_2024} and work by \citet{yatim_space-time_2024}  have extended this idea to the video space, using diffusion features to enable prompt-based video editing text-driven motion transfer.

\subsubsection{A general representation extraction framework}

\begin{figure*}
    \centering
    \includegraphics[scale=0.4]{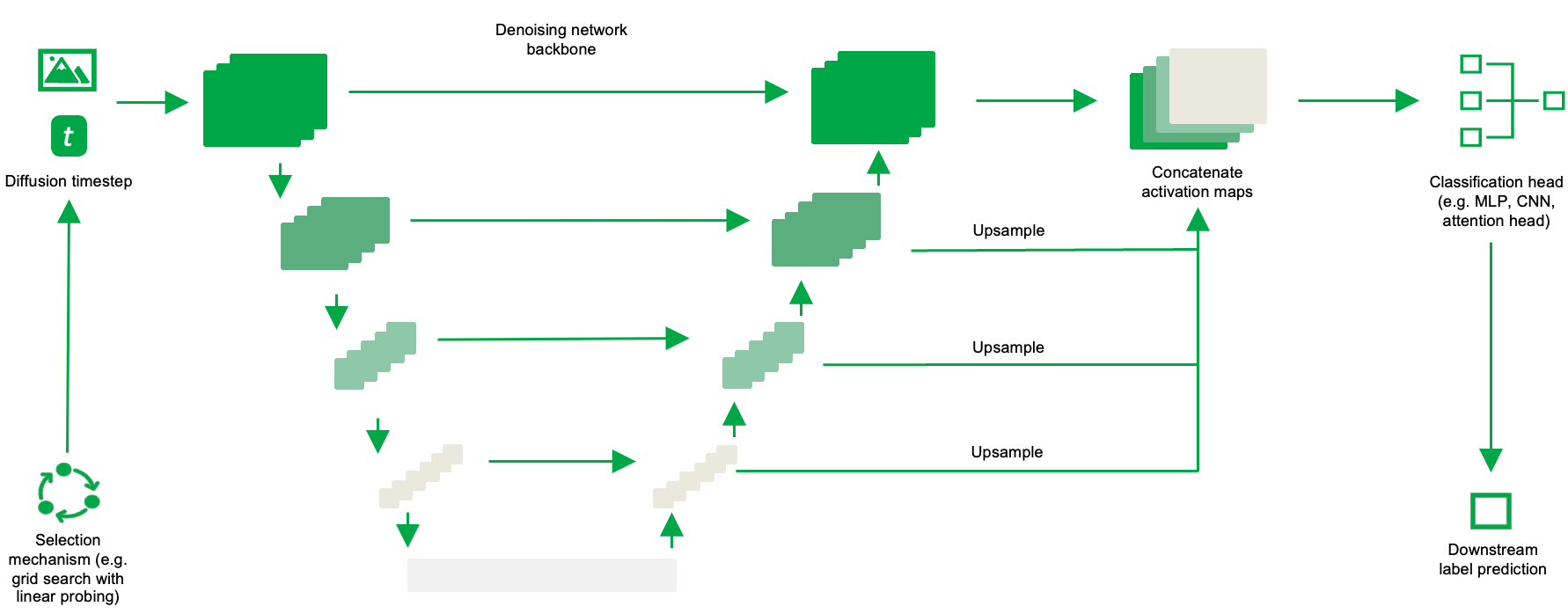}
    \caption{A high-level overview of a framework for extracting representations from pre-trained diffusion models for downstream tasks.}
    \label{fig:framework1}
\end{figure*}

Many of the methods outlined in the previous section follow a similar procedure in leveraging learned representations of pre-trained diffusion models for downstream vision tasks. In this section, we aim to consolidate these approaches to a common three-step framework. We do this to provide clarity on the relationship between diffusion models and their use for downstream predictive tasks. To leverage intermediate activations for downstream tasks, a selection methodology that outputs the ideal diffusion timestep input as well as the intermediate layer number(s) whose activation maps have the highest predictive performance when upsampled and linearly probed must be applied. This can be a trainable model \cite{luo_diffusion_2024}, a grid search procedure \cite{xiang_denoising_2023} or a learning agent \cite{yang_diffusion_2023}. The goal of this methodology is generally to select timestep $t \in T$ and a set of decoder block numbers $B$ that maximize predictive performance on a downstream task. Given a set of possible timesteps $T$ and a set of decoder blocks $\mathcal{B}$, the goal is to find:
\begin{equation}
(t^*, B^*) = \arg\min_{t \in T, B \subseteq \mathcal{B}} \mathcal{L}_{\text{discr}}(t, B)
\end{equation}

where $\mathcal{L}_{\text{discr}}(t, B)$ represents the discriminative loss at timestep $t$ when the blocks in $B$ are used for downstream prediction. Generally, discriminative tasks will require more high-level features corresponding to structural elements and shapes, whereas generative tasks mapping random noise to images will require the computation of lower-level features. The ideal intermediate layer number as well as the optimal diffusion timestep will largely depend on the exact downstream prediction task, the dataset, and the architecture of the diffusion model used. 
\par
Once the ideal timestep and layer number are determined, an input image and the selected diffusion timestep are passed to the diffusion model, and the intermediate activations in the selected decoder blocks computed in the forward pass are extracted and generally concatenated and pre-processed depending on the downstream task (e.g. through upsampling, pooling, etc.). Finally, a classification head is trained on the annotated dataset, taking the preprocessed features extracted from the diffusion model as input. This classification head can be an MLP, a CNN, or an attention-based network depending on the availability of labeled data and predictive performance on the dataset. The diffusion model weights are usually frozen in this probing process, but additional fine-tuning regimes can increase discriminative performance for certain datasets and architectures (see e.g., \citet{xiang_denoising_2023}). Fig.~\ref{fig:framework1} shows an overview of the generalized framework.

\subsubsection{Knowledge transfer}

Aside from leveraging intermediate activations from pre-trained diffusion models directly as inputs to a recognition network, several recent approaches propose a more indirect method of reusing learned representations for downstream tasks. We summarize these under the term \textit{knowledge transfer} methods. This reflects the common idea of distilling representations from pre-trained diffusion models and then transferring them to auxiliary networks in a way that is distinct from simply providing aggregated feature activation maps as input. Several of these approaches are discussed in the following section.
\par
\citet{yang_diffusion_2023} propose RepFusion, a knowledge distillation approach that dynamically extracts intermediate representations at different time steps using a reinforcement learning framework, and uses the extracted representations as auxiliary supervision for student networks. Given an input $\mathbf{x}$ with label $\mathbf{y}$, the authors extract a pair of features, one from the diffusion probabilistic model (DPM) and one from the student model, where $\mathbf{z^{(t)}}$ is the diffusion model representation and $\mathbf{z}$ is the student model representation. The distance between the two is minimized during training using a loss function $\mathcal{L}_{kd}$. After the distillation, the student network is reapplied as a feature extractor and fine-tuned on the available task labels. Previous approaches for using diffusion model representations rely on grid-search to determine which diffusion timestep to use for feature extraction. Here, the authors formulate a reinforcement learning environment where the action space is the set of all possible timesteps $t$ available for selection, and the reward function is the negative task loss $-\mathcal{L}_{task}(\mathbf{y}, g(\mathbf{z}^{(t)};\theta_g))$. Given the input $\mathbf{x}$, a policy network $\pi_{\theta_{\pi}}(t | \mathbf{x})$ is trained to determine which timestep $t$ to use for representation extraction. Once the timestep is selected, the authors use the feature representations in the mid-block of the DPM for the selected timestep $t^*$ to obtain $\mathbf{z}^{(t^*)}$. After the distillation phase, the student network is used as a feature extractor and subsequently fine-tuned on the task label $\mathbf{y}$.
\par
\citet{li_dreamteacher_2023} introduce DreamTeacher, a knowledge distillation method using a feature regressor module that distills the learned representations of a generative model $G$ into a target image recognition backbone $f$. Given a feature dataset $D = \{\mathbf{x}_i, \mathbf{f}_i^g\}^N_{i=1}$ consisting of images $\mathbf{x}$ and extracted features $\mathbf{f}_i^g$, $f$ is trained by distilling $\mathbf{f}_i^g$ into the intermediate features of $f(\mathbf{x}_i)$. The features are extracted from $G$ by running a forward diffusion process for $T$ timesteps and conducting a single denoising step to extract $\mathbf{f}_i^g$ from the intermediate layers of the U-Net backbone. The extracted features are distilled using a feature regressor module with a top-down architecture containing lateral skip connections that aligns the image backbone features with the generative features. Intermediate CNN encoder features $\mathbf{f}^e_l$ at layers $l$ and regressor outputs $\mathbf{f}^r_l$ are used to compute an MSE feature regression loss inspired by FitNet \cite{romero_fitnets_2015}:
\begin{equation}
    \mathcal{L}_{\text{MSE}} = \frac{1}{L} \sum_{l=1}^{L} \left\| f_{l}^{r} - \mathcal{W}(f_{l}^{e}) \right\|_{2}^{2}
\end{equation}

$\mathcal{W}$ is a non-learnable operator implemented as LayerNorm \cite{ba_layer_2016}. This loss is combined with the activation-based Attention Transfer (AT) objective \cite{zagoruyko_paying_2017}, which distills a one dimensional "attention map" for each spatial feature. DreamTeacher is evaluated on a range of downstream recognition tasks by fine-tuning the pre-trained backbone with additional classification heads for each task. DreamTeacher outperforms existing contrastive and masking-based self-supervised methods on the COCO \cite{lin2014microsoft_mscoco}, ADE20k \cite{zhou_semantic_2019} and BDD100K \cite{yu_bdd100k_2020} benchmarks.
\par
Both RepFusion and DreamTeacher are inspired by earlier works on knowledge distillation \cite{hinton_distilling_2015, romero_fitnets_2015}. \citet{li_your_2023} propose a slightly different knowledge transfer approach: Diffusion Classifier, a method for zero-shot classification that leverages conditional density estimates from text-to-image diffusion models. This classifier converts the diffusion model into a classifier by computing class conditional likelihoods $p_{\theta}(x | \mathbf{c}_i)$ and using Bayes' theorem to obtain predicted class probabilities $p(\mathbf{c}_i | \mathbf{x})$. Since direct computation of $p_{\theta}(x | \mathbf{c}_i)$ is intractable, they use the Evidence Lower Bound (ELBO) in its place. The classifier is derived by adding noise repeatedly and estimating noise reconstruction losses for each class using Monte Carlo methods. While Diffusion Classifier suffers from high inference time, it generally outperforms DDPM-Seg \citet{baranchuk_label-efficient_2022} on most datasets and is competitive with CLIP ResNet-50 \cite{radford2021learning_clip} and OpenCLIP ViT-H/14 \cite{cherti_reproducible_2023}.
\par

\subsubsection{Reconstructing diffusion models}

Previous diffusion representation learning techniques do not propose making fundamental modifications to diffusion model architectures and training methodologies. While these techniques often show encouraging performance for downstream tasks, they fail to generate deep insights into the architectural components and techniques required to learn useful representations. It remains largely unclear for example whether the representation learning abilities of diffusion models are driven by the diffusion process, or by the model's denoising capabilities. It is also unclear what architectural and optimization choices can improve diffusion models' representation learning capabilities.
\par
\citet{chen_deconstructing_2024} investigate these questions by deconstructing a denoising diffusion model (DDM), modifying individual model components to turn a DDM into a Denoising Autoencoder. The deconstruction process consists of three stages. In the first stage, the DDM is reoriented for self-supervised learning. This entails the removal of class conditioning and a reconstruction of the VQGAN tokenizer~\cite{esser_taming_2021} used in the DiT baseline. Both the perceptual and adversarial loss terms rely on annotated data and are thus removed. This essentially converts the VQGAN to a VAE. The second stage consists of simplifying the VAE tokenizer even further, replacing it with different autoencoder variants. Surprisingly, the authors find that using simpler autoencoder variants, like patch-wise PCA, does not degrade performance substantially. The authors conclude that the dimensionality per token of the latent space has a much larger impact on probing accuracy than the chosen autoencoder. The final deconstruction step includes converting the DDM to predict the denoised input instead of the added noise and removing input scaling, as well as changing the diffusion model to operate directly in the pixel space. This final stage results in what the authors call the latent Denoising Autoencoder (l-DAE). They conclude that representation learning abilities are largely driven by the denoising-driven process rather than the diffusion process.
\par
l-DAE is inspired by the observation that diffusion models resemble hierarchical autoencoders with varying noise scales. This insight is also applied in DiffAE \citep{preechakul2022diffusion_autoencoder}, which uses diffusion models for representation learning via autoencoding. \citet{preechakul2022diffusion_autoencoder} separate latent representations into a compact semantic representation and a stochastic representation. DiffAE consists of a semantic encoder, that generates a semantic representation $\mathbf{z}_{\text{sem}}$, as well as a conditional DDIM \cite{song2020denoising_ddim}. This DDIM acts both as the stochastic encoder, which maps $\mathbf{x}_0$ to $\mathbf{x}_T$, and as the decoder, which maps $\mathbf{x}_T$ to $\mathbf{x}_0$. $\mathbf{x}_T$ represents the stochastic representation and captures low-level variation, whereas $\mathbf{z}_{\text{sem}}$ encodes higher-level semantics. During inference, \citep{preechakul2022diffusion_autoencoder} fit a second latent DDIM to $\mathbf{z}_{\text{sem}}$, and sample from this DDIM and $\mathbf{x}_T$ to facilitate unconditional sampling. Variations in $\mathbf{x}_T$ with fixed $\mathbf{z}_{\text{sem}}$ result in minor changes in generated images, while varying $\mathbf{z}$ leads to different reconstructions, showing DiffAE's efficiency in generating semantically meaningful and decodable representations. InfoDiffusion \cite{wang_infodiffusion_2023} extends DiffAE, supporting custom priors and improving latent representations $\mathbf{z}_{\text{sem}}$ via mutual information regularization.
\par
\citet{zhang_unsupervised_2022} observe that there is a gap between the true and the predicted posterior mean of $\mathbf{x}_{t-1}$ when predicting from $\mathbf{x}_t$ in the diffusion reverse process. Classifier guidance can be viewed as reconstructing information lost in the diffusion forward process by shifting the posterior mean to fill that gap. They propose Pre-trained DPM AutoEncoding (PDAE), a method for adapting DPMs to decoders for image reconstruction. Instead of using a class label $\mathbf{y}$ to fill this information gap, PDAE employs a model to predict mean shift according to encoded representations $\mathbf{z}$, ensuring that $\mathbf{z}$ contains as much information as possible from $\mathbf{x}_0$. Specifically, \citet{zhang_unsupervised_2022} employ an encoder $E_{phi}(\mathbf{x}_0) = \mathbf{z}$ along with a gradient estimator $G_{\psi}(\mathbf{x}_t, \mathbf{z}, t)$ that simulates $\nabla_{\mathbf{x}_t} \log(p(\mathbf{z}|\mathbf{x}_t)$ to modify the conditional DPM training objective. This modified objective forces the predicted mean shift to fill the aforementioned posterior mean gap. With a trained $G_{\psi}(\mathbf{x}_t, \mathbf{z}, t)$, the score of the implicit classifier $p(\mathbf{z}|\mathbf{x}_t)$ can be used analogously to classifier-guided sampling. PDAE is evaluated using similar experiments as used in \cite{preechakul2022diffusion_autoencoder} and exhibits improved training efficiency and performance.
\par
\citet{pan_masked_2024} propose a different method for DDM reconstruction. They introduce a masked diffusion model (MDM), designed for self-supervised semantic segmentation. MDM substitutes the conventional diffusion process with a masking mechanism inspired by the masked autoencoder \citep{he2022masked_mae}. The representations learned by the pre-trained MDM are extracted following \citet{baranchuk_label-efficient_2022}. The proposed MDM is a variant of a time-dependent denoising autoencoder, that takes a masked input image and subsequently reconstructs the uncorrupted image. While other DDMs and MAE use an MSE reconstruction loss, \citet{pan_masked_2024} propose using the structural similarity index (SSIM) loss. This is done to narrow the gap between reconstruction and subsequent segmentation tasks. MDM is pre-trained on a set of unlabeled images using the described self-supervised approach. The learned representations are then extracted to train an MLP-based classification head on a smaller labeled dataset. Features based on specific block setting $\mathcal{B}$ are extracted by selecting the activation maps from each of the specified blocks, upsampling activation maps to match the image size, and concatenating the activations. The method achieves state-of-the-art results against existing supervised segmentation methods on multiple benchmark datasets even when only 10\% of labels are available. DiffMAE \cite{wei_diffusion_2023} is a similar approach that uses a conditional generative objective, where the distribution of the masked pixels $\mathbf{x}_0^m$ conditioned on the visible pixels $\mathbf{x}_0^v$ is modeled, and diffusion is only applied to masked regions.
\par
\par
\citet{hudson_soda_2024} introduce a novel view generation learning goal as well as a bottleneck layer to aid representation learning. They present SODA, a self-supervised diffusion model that consists of an encoder and a denoising decoder. The encoder produces a concise latent representation, which is used for denoising decoder guidance by modulation of the decoder activations. The encoder $E(\mathbf{x})$ converts an input view $\mathbf{x}$ into a compressed latent representation $\mathbf{z}$, which is used to generate a novel output view $\mathbf{x}'$ relating to the input $\mathbf{x}$. $\mathbf{x}'$ is created through a diffusion process conditioned on the latent representation $\mathbf{z}$ via feature modulation. In addition to this, the authors use layer modulation, where the latent representation is partitioned, with each partition $\mathbf{z}_i$ modulating a specific pair of layer activations. This enables further specialization among the latent subvectors, where some are optimized to capture finer levels of granularity than others. During training, \citet{hudson_soda_2024} opt to randomly zero out a subset of the latent subvectors, effectively implementing a layer-wise generalization of classifier-free guidance. This further increases control over the generative process since the trained model can then be conditioned using a curated subset of latent subvectors.
\par
SmoothDiffusion \cite{guo_smooth_2024} is a work focusing on improving the smoothness of the latent space of diffusion models, which refers to the consistency of perturbations in the latent and the image space. SmoothDiffusion enforces smoothness over its latent space by proposing a novel step-wise variation regularization method in training. The resulting smoothed latents benefit a wide range of image interpolation, image inversion and image editing tasks.

\subsubsection{Joint diffusion models}

Many current diffusion-based representation learning methods focus on using the diffusion model's latent variables to benefit the training of a separate recognition network. These frameworks are conceptually equivalent to constructing hybrid models that solely concentrate on synthesis in the pre-training stage, and on downstream recognition in the post-training/fine-tuning phase. The recognition head and the diffusion denoising network do not share a parametrization, and the recognition head is often trained separately while keeping the weights of the denoising network frozen. A natural question that arises is whether this separation is necessary and whether approaches that optimize a generative and a discriminative objective simultaneously in a shared parametrization can improve representation learning.
\par
HybViT \cite{yang_your_2022} is an approach that establishes a direct connection between diffusion models and vision transformers by training a single hybrid model for both image classification and image generation. This hybrid model uses a shared parametrization for image classification and reconstruction. The authors use a ViT backbone to train a model with a combined loss $\mathcal{L}$ consisting of a standard cross-entropy loss to train $p(y | \mathbf{x}$) and the simple denoising loss to train $p(x)$. HybViT provides stable training and outperforms previous hybrid models on both generative and discriminative tasks, but lags behind generative-only models in generation quality. HybViT also requires more training iterations to achieve high classification performance, and the sampling speed during inference is slow.
\par
Joint Diffusion Models (JDM) \cite{deja_learning_2023} is a related work that produces meaningful representations across generative and discriminative tasks. Using a U-Net backbone, JDM consists of an encoder $e_{\nu}$, a decoder $d_{\psi}$, and a classifier $g_{\omega}$. The encoder maps an input $\mathbf{x}_t$ to feature vectors $\mathbf{Z_t} = e_{\nu}(\mathbf{x}_t)$. The decoder reconstructs these into a denoised sample $\mathbf{x}_{t-1} = d_{\psi}(\mathbf{Z}_t)$, and the classifier predicts the target class $\hat{y} = g_{\omega}(\mathbf{Z}_t)$. The combined training objective includes cross-entropy loss $L_{\text{class}}$ and the noise prediction network's simplified objective $L_{t, \text{diff}}(\nu, \psi)$, resulting in the following loss:
$$
L(\nu, \psi, \omega) = L_{\text{class}}(\nu, \omega) - L_0(\nu, \psi) - \sum_{t=2}^{T} L_{t, \text{diff}}(\nu, \psi) - L_T(\nu, \psi).
$$

JDM also enables a simplification of classifier guidance. By applying the classifier to noisy images $\mathbf{x}_t$, the classifier is effectively augmented to be robust to noise. To guide the generated sample towards a target label, representations $\mathbf{Z}_t$ are optimized according to the classifier gradient, giving $\mathbf{Z}_t' = \mathbf{Z}_t - \alpha \nabla_{\mathbf{z}_t} \log g_{\omega}(\mathbf{y} | \mathbf{Z}_t)$. JDM achieves state-of-the-art performance for joint models on CIFAR and CelebA datasets, outperforming HybViT.
\par
\citet{tian_addp_2024} propose the Alternating Denoising Diffusion Process (ADDP). ADDP alternately denoises pixels and VQ tokens. Given an image $\mathbf{x}_0$, a pre-trained VQ Encoder \cite{chang_maskgit_2022} maps time image to VQ tokens $\mathbf{z}_0$. The alternating diffusion process masks regions of $\mathbf{z}_0$ with a Markov chain according to diffusion timestep $t$, producing $\mathbf{z}_t$. Unreliable tokens $\Bar{\mathbf{z}}_t$ are generated by a token predictor and fed into a VQ Decoder to synthesize $\mathbf{x}_t$, replacing the masked regions of $\mathbf{z}_0$. A pixel-to-token generation network is then trained to approximate the distribution of $\Bar{\mathbf{z}}_{t-1}$. During sampling, ADDP starts with a representation of pure unreliable tokens $\Bar{\mathbf{z}}_T$ and iteratively denoises the token sequence by predicting $\Bar{\mathbf{z}}_{t-1}$. For recognition, the representations learned by the pixel-to-token generation network can be forwarded to different task-specific recognition heads. ADDP with the VQGAN tokenizer \cite{esser_taming_2021} MAGE-Large \cite{li_mage_2023} token predictor and ViT-Large \cite{dosovitskiy2020image_vit} pixel-to-token encoder, outperforms previous unified models in image classification, object detection, semantic segmentation, and unconditional generation.

\subsubsection{Generative augmentation}

A lot of state-of-the-art representation learning methods \cite{he2020momentum_moco, grill_bootstrap_2020, chen_simple_2020} rely on a fixed set of data augmentations to define positive labels for learning representations. This approach encourages encoders to learn to map the original and the augmented image to similar embedding space representations \cite{ayromlou_can_2024}. These augmentations should not alter the semantics of the image, and they should not render the image unrealistic in a real-world setting. A set of standard transformations might not adequately capture the distribution of real-world data, raising the question of how to design transformations that create diverse images and improve the generalization of learned representations.
\par
\citet{ayromlou_can_2024} propose using latent diffusion models \cite{rombach2022high_latentdiffusion_ldm} to generate novel views of the original image that preserve the semantic content, while closely following the distribution of real images. This augmentation method is denoted by:
\begin{equation}
T_0(\mathbf{x}) =
\begin{cases}
G(\mathbf{z}; \phi(\mathbf{x})) & \text{if } p \leq p_0 \\
\mathbf{x} & \text{otherwise},
\end{cases}
\end{equation}

where $G$ denotes a conditional generative model taking noise vector $\mathbf{z} \sim  \mathcal{N}(0, \mathbf{I})$ and condition vector $\phi(\mathbf{x})$ as inputs. $\phi$ is a pre-trained image encoder such as CLIP \cite{radford2021learning_clip}, $p \in [0, 1]$ is a random number and $p_0$ is a hyperparameter specifying the probability of applying the augmentation. \citet{ayromlou_can_2024} show that using generative augmentation leads to consistent improvements in learned representations over standard transformations across other representation learning techniques.
\par
\citet{shipard2023diversity} take this approach one step further, using Stable Diffusion to generate a fully synthetic dataset to improve model-agnostic zero-shot classification (MA-ZSC). They use Stable Diffusion, employing several variations of prompts designed to increase the diversity of the synthetic dataset. An image classifier is subsequently trained on this synthetic dataset, and zero-shot classification results on CIFAR10, CIFAR100, and EuroSAT \cite{helber_eurosat_2019} are evaluated. \citet{shipard2023diversity} observe substantial classification architecture-agnostic improvements on the aforementioned datasets, achieving comparable performance to state-of-the-art zero-shot classification methods like CLIP.
\par
Moving beyond classification, \citet{schnell_scribblegen_2024} apply similar ideas to scribble-supervised segmentation \cite{lin_scribblesup_2016, pan2021scribble}, a weakly-supervised form of semantic segmentation that uses sparse annotations in the form of scribbles drawn over the images. They introduce ScribbleGen, a diffusion model conditioned on semantic scribbles that generates synthetic training images for data augmentation. ScribbleGen utilizes a ControlNet \cite{zhang_adding_2023} denoising diffusion model for noise prediction given $\mathbf{x}_t$ and conditioning signal $\mathbf{c}$. The number of classes is denoted by different color scribbles in RGB images, and the conditioning signal $c$ is supplemented by a text prompt stating all classes in the image. \citet{schnell_scribblegen_2024} trade-off photorealism and image diversity by introducing an encode ratio $\lambda \in [0, 1]$. This diffusion parameter controls the number of noise-adding forward diffusion steps, where $\lambda = 1$ leads to no change but $\lambda < 1$ leads to $\lambda \cdot T$ steps, meaning less noise is added to the input image. The authors evaluate both a fixed and an adaptive $\lambda$, where the encoding ratio is gradually increased to provide increasingly diverse synthetic images during training. ScribbleGen achieves state-of-the-art performance on the PASCAL VOC12 segmentation dataset \cite{everingham_pascal_2010} using scribbles from Scribblesup \cite{lin_scribblesup_2016}.
\par
DiffuMask \cite{wu_diffumask_2023} is another generative augmentation method designed to improve downstream semantic segmentation tasks. The idea here is to exploit cross-attention maps between text prompts and generated images to extend image synthesis to semantic mask generation. Synthetically generated masks are used for data augmentation to improve downstream segmentation performance. Individual token attention maps of all layers are averaged and converted to binary masks using an adaptive threshold mechanism based on an AffinityNet \cite{ahn_learning_2018}. Additionally, a noise-learning module prunes low-quality segmentation masks, and the authors employ several prompt engineering and static image transformations to further enhance the diversity of the generated images and corresponding segmentation masks.

\subsection{Representation Learning for Diffusion Model Guidance}
\label{subsec:r4d}

\begin{figure*}
    \centering
    \includegraphics[width = \textwidth]{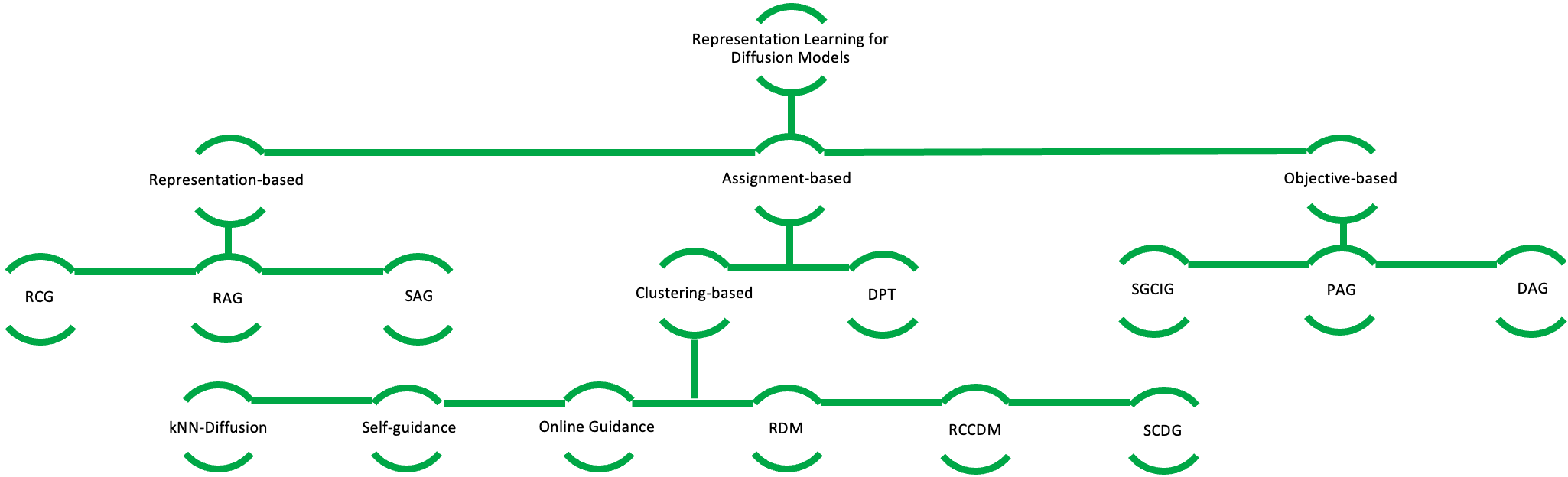}
    \caption{A hierarchical overview of current diffusion model training frameworks that leverage representation learning techniques for conditional generation and guidance.}
    \label{fig:hierarchy}
\end{figure*}

Despite the remarkable performance of generative models, there exists a gap in quality between conditional and unconditional image generation approaches \cite{icgan}. This is especially the case for GANs \cite{goodfellow_generative_2014}, which suffer from mode collapse when trained in a fully unsupervised setting \cite{liu_diverse_2020}. Unconditional GANs often fail to accurately model multi-modal distributions, e.g. not being able to generate all digits for MNIST \cite{liu_diverse_2020}. Class-conditional GANs \cite{brock2018large_biggan} \cite{mirza2014conditional_cgan} mitigate this issue, but require labeled data. Recent approaches like self-conditioned GANs \cite{liu_diverse_2020} and instance-conditioned GANs \cite{icgan} attempt to train conditional GANs without requiring labeled data, and are able to achieve competitive generation results.
\par
Diffusion models have since surpassed the image generation capabilities of GANs \cite{dhariwal2021diffusion_beat}, but suffer from a similar performance discrepancy between conditional and fully self-supervised approaches. Current state-of-the-art diffusion models are conditional models that rely on guidance approaches that also require annotated data. Self-supervised guidance approaches can leverage much larger unlabeled datasets for pre-training, and thus have the potential to transcend current image generation approaches. One intuitive approach for leveraging representation learning to facilitate these guidance methods is to explore methods that assign labels to unlabeled data, e.g. through clustering and classification approaches. We introduce several approaches in the following section. Fig.~\ref{fig:hierarchy} shows a proposed taxonomy of representation learning techniques for diffusion guidance.

\subsubsection{Assignment-based guidance}

\citet{ashual2022knn} propose kNN-Diffusion, an efficient text-to-image diffusion model trained without large-scale image text pairings. To facilitate text-guided image generation without paired text-image data, a shared text-image encoder mapping text-image pairs into the same latent space is required. The authors use CLIP to achieve this, a pre-trained encoder trained using contrastive loss on a large-scale text-image pair dataset. kNN-Diffusion leverages $k$-Nearest-Neighbors search to generate $k$ embeddings from a retrieval model. The retrieval model uses the input image representation during training, and the text prompt representation curing inference. This approach eliminates the need for annotated data but still requires a pre-trained encoder like CLIP, which in turn requires a large-scale dataset of text-image embeddings for pre-training.
\par
\citet{blattmann2022retrieval} propose retrieval-augmented diffusion models (RDM), which equip diffusion models with an image database for composing new scenes based on retrieved images. Inspired by advances in retrieval-augmented NLP \cite{borgeaud_improving_2022,wu_memorizing_2022}, RDM enhances performance with fewer parameters and computational resources. Despite being trained only on images, RDM allows conditional synthesis due to the shared image-text feature space of CLIP \cite{radford2021learning_clip}. RDM includes a trainable conditional latent diffusion model $p_{\theta}$, an external image database $\mathcal{D}$, and a fixed sampling strategy $\xi_k$ that selects a subset $\mathcal{M}_{\mathcal{D}}^{(k)}$ of $\mathcal{D}$ based on a query $\mathbf{x}$. One strategy $\xi_k(\mathbf{x}, \mathcal{D})$ is to retrieve the $k$ nearest neighbors using a distance function $d(\cdot, \mathbf{x})$. The retrieved data is processed through a frozen image encoder $\phi$ and used to condition $p_{\theta}$. During training, $\xi_k$ retrieves $k$ nearest neighbors for a query image $\mathbf{x}$ using cosine similarity in CLIP's image feature space as the distance function $d(\mathbf{x},\mathbf{y})$. This approach ensures that retrieved image representations are useful for generation tasks and allows for text conditioning due to CLIP's shared feature space. The dataset $\mathcal{D}$ and retrieval strategy $\xi_k$ can be changed at test time, adding flexibility for different conditioning modalities and adaptability to other data distributions.
\par
\citet{hu_self-guided_2023} propose a method also motivated by eliminating the need for annotated data. Self-guided diffusion is a framework encompassing a feature extraction function $g_{\phi}$ and a self-annotation function $f_{\psi}$. The feature extraction function is a self-supervised feature extractor that maps the input data $\mathbf{x} \in \mathcal{D}$ to a feature space $\mathcal{H}$, where $\mathcal{D}$ denotes the dataset. This feature representation is an input of $f_{\psi}$, which maps feature representation $g_{\phi}(\mathbf{x}; \mathcal{D})$ to a guidance signal $k$. This framework can be applied to achieve self-labeled guidance, where $k$ is a one-hot embedding derived using $k$-means clustering as the self-annotation function $f$ on compacted features generated by $g_{\phi}$. More fine-grained spatial guidance is achieved by self-boxed guidance, which uses a mapping from feature space $\mathcal{H}$ to a bounding box as the self-annotation function $f$, as well as self-segmented guidance, which uses a mapping to a segmentation mask to generate guidance signals by clustering. Self-guidance significantly outperforms unconditional diffusion models, and even outperforms classifier-free guided diffusion models that use ground-truth annotations on image generation. This suggests that the clusters are potentially more aligned with the visual similarity of the images, and are better guidance signals than ground-truth labels alone. While this approach is self-supervised, it still relies on an external pre-trained feature extractor to generate feature representations for clustering.
\par
For this reason, \citet{hu_guided_2023} extend their work to propose an online feature clustering method using the Sinkhorn-Knopp algorithm. This is challenging since the idea requires obtaining conditioning signals for clustering during training from a diffusion model that is dependent on this conditioning. This issue is solved by introducing a zero vector into the conditional diffusion model for the signals used to identify the clustering. For each image example, the conditional diffusion model conditioned on this zero vector undergoes a fully-connected feature prediction head used to compute features that are mapped to a set of learnable prototypes denoted $M$. This method uses a combination of the diffusion training loss and a Sinkhorn-Knopp loss to achieve guidance signals $\mathbf{c}$ that are based on clustering features using $M$. The promise of this method is high, with self-guided diffusion outperforming related unconditional generation baseline comparisons on ImageNet256 and LSUN-Churches while being competitive with class guidance methods that rely on ground truth labels. The online approach specifically does not rely on ground truth labels or any external pre-trained models. \citet{adaloglou_rethinking_2024} build on the aforementioned cluster-based guidance approaches by utilizing EDM \cite{karras2022elucidating}, TEMI clustering \cite{adaloglou_exploring_2023} and a method for deriving an upper cluster bound for feature-based clustering. 
\par
Other approaches to diffusion model guidance rely on generating pseudo-labels for unlabeled data. \citet{you_diffusion_2023} propose dual pseudo training (DPT), which uses a classifier trained on limited labeled data to generate pseudo-labels. These are then used to condition a diffusion model to generate pseudo images, which are in turn used as data augmentation to retrain a classifier on a mix of pseudo and real images. DPT involves three stages. First, a semi-supervised classifier is trained on partially labeled data to predict pseudo-labels $\hat{\mathbf{y}}$ for all images $\mathbf{x} \in \mathcal{X}$. Second, a conditional generative model is trained on the dataset $S_1 = \{(\mathbf{x}, \mathbf{y}) | \mathbf{x} \in \mathcal{X}\}$ with pseudo-labels. Finally, the classifier is retrained on real data that is augmented by the generated data. DPT achieves highly competitive performance on ImageNet classification and generation with as little as five labels per class, outperforming several supervised diffusion model benchmarks like ADM~\cite{dhariwal2021diffusion_beat} and LDM~\cite{rombach2022high_latentdiffusion_ldm}.

\subsubsection{A generalized framework for assignment-based guidance}

\begin{figure*}
    \centering
    \includegraphics[width = \linewidth]{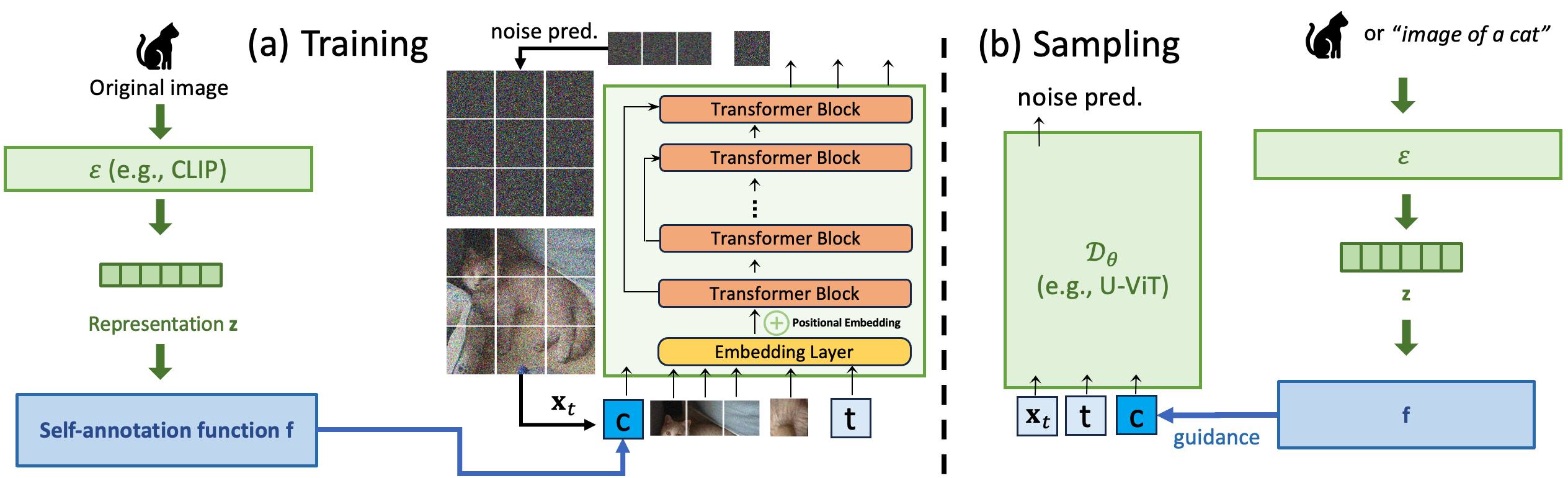}
    \caption{A generalization of assignment-based guidance training and sampling pipelines. Samples are conditioned on annotations generated by a self-annotation function $f$, using features extracted by a pre-trained image encoder (e.g., CLIP \cite{radford2021learning_clip}).}
    \label{fig:framework2}
\end{figure*}

Assignment-based guidance approaches all rely on assigning annotation to inputs during training, which enables controlled generation during inference when conditioning on this annotation. We therefore propose to formulate a generalized framework that encapsulates all assignment-based guidance approaches discussed here. This framework consists of three main components. The first is a self-supervised image encoder $\mathcal{E}(\mathbf{x})$, that maps inputs to a low-dimensional feature representation $\mathbf{z}$. Using a multi-modal feature extractor like CLIP has the advantage of enabling text-based as well as image-based conditioning, but other feature extractors can be used, provided they generate semantically meaningful image representations.
\par
The second is a self-annotation function $f(\mathbf{z})$, which uses the image representation to produce annotation $\mathbf{c}$ for input image $\mathbf{x}$. In the simplest case, this self-annotation function is an external pre-trained image classifier that generates pseudo-class labels from image representations, similar to the approach employed in DPT \cite{you_diffusion_2023}, where the external classifier is subsequently re-trained on the conditionally generated images. In other cases, the self-annotation function is a retrieval model, which uses a distance function $d$ to retrieve images similar to the training image, and uses representations of the retrieved images for generating the guidance signal $\mathbf{c}$.
\par
The final component is a denoising network $\mathcal{D}_{\theta}(\mathbf{x}_t, \mathbf{c}, t)$, which takes the noisy image $\mathbf{x}_t$, the diffusion timestep $t$ and the guidance signal $\mathbf{c}$ as input, and denoises the image. During inference, controlled generation is enabled by passing an initial guidance signal $\mathbf{k}$ (which can be multi-modal as long as the embedding space of the encoder $\mathcal{E}$ is shared between modalities) through the encoder to generate representation $\mathbf{z} = \mathcal{E}(\mathbf{k})$. The conditioning signal $\mathbf{c}$ is then generated by passing $\mathbf{z}$ to the self-annotation function $f$ where $\mathbf{c} = f(\mathbf{z})$. Passing $\mathbf{x}_t$, $\mathbf{c}$ and $t$ to the denoising network $\mathcal{D}_\theta$ now enables synthesis of novel images semantically similar to the initial guidance signal $\mathbf{k}$.
\par
One of the main motivations behind the design of assignment-based guidance methods is the reliance on existing methods on labeled data. While it could be argued that the aforementioned assignment-based guidance approaches are indirectly reliant on annotated data through the pre-trained image encoder, it is important to note that this encoder can be replaced with a fully self-supervised encoder as well. CLIP relies on the availability of a large-scale dataset of image-caption pairs and is thus not fully self-supervised, but other representation learning methods are also able to generate semantic representations. CLIP is used in many approaches to facilitate both text prompt-based and image conditioning during inference, which may no longer be possible when using primarily image-based feature extractors. A summary of the training and inference methodology can be found in Fig.~\ref{fig:framework2}.

\subsubsection{Representation-based guidance}

\citet{li_return_2024} present Representation-Conditioned Image Generation (RCG), a framework conditioning diffusion models on a self-supervised representation distribution mapped from the image distribution using a pre-trained encoder. The idea is to train a Representation Diffusion Model (RDM) on the representations generated by a pre-trained encoder to generate low-dimensional image representations. After this, a pixel generator conditioned on the representation is trained to map noise distributions to image distributions. RCG consists of three main components. The first is a pre-trained image encoder, which converts the original image distribution into a representation distribution. The authors propose using self-supervised contrastive learning methods (e.g. MoCo v3) for generating this representation distribution. The second is a representation generator in the form of an RDM, which learns to generate representations from Gaussian noise following the DDIM \cite{song2020denoising_ddim} sampling process. The final component is a pixel generator that crafts image pixels conditioned on image representations. RCG can easily incorporate classifier-free guidance for unconditional generation tasks, since the pixel generator is conditioned on self-supervised representations. RCG emerges as a highly promising method for bridging the gap between conditional and unconditional image generation, outperforming pre-existing unconditional generation approaches on ImageNet, and exhibiting competitive performance with current state-of-the-art class-conditional approaches.
\par
Readout Guidance (RG) \cite{luo_readout_2024} makes use of auxiliary readout heads trained on top of a frozen diffusion model to extract properties of the generated image that can be used for guidance. These properties can include human pose, depth maps, edges, and even higher-order properties like similarity to another image. During sampling, the properties extracted by the readout heads can be compared to user-defined control targets, and used in a methodology similar to classifier guidance \cite{dhariwal_diffusion_2021} to guide generation.
\par
\citet{lin_diffusion_2024} identified a novel self-perceptual objective that enhances diffusion models, enabling them to generate more realistic samples. Contrary to the conventional approach of training or employing an image encoder, the authors demonstrate that a pre-trained diffusion model inherently functions as a perceptual network and can be used to generate perceptual representations. The perceptual loss facilitates the model's ability to generate more realistic images even with unconditional synthesis.
\par
Also inspired by the downsides of classifier guidance and classifier-free guidance, \citet{hong_improving_2023} introduce Self-Attention Guidance (SAG). SAG adversarially blurs regions that contain salient information by leveraging intermediate self-attention activation maps, using the residual information as guidance. This increases the generation quality without requiring external information or additional training. The self-attention mechanism, contained in both U-Net and DiT diffusion backbones, allows the noise predictor to attend to the most informative features of the input. The self-attention maps $\mathbf{A}_t^S \in \mathrm{R}^{N \times (HW) \times (HW)}$ are aggregated and reshaped to dimension $\mathrm{R}^{H \times W}$ using global average pooling and nearest-neighbor upsampling to match the resolution of $\mathbf{x_t}$. The difference between the blurred image $\mathbf{\Tilde{\mathbf{x}}_t}$ and $\mathbf{x}_t$ is used as conditioning, thereby retaining the information masked in this process.

\subsubsection{Objective-based guidance}

Many of the previous outlined approaches focus on eliminating the need for pre-trained classifiers, encoders and dataset annotations for training conditional diffusion models. Other recent works \cite{kim_depth-aware_2024,epstein_diffusion_2023} have demonstrated that internal diffusion model representations can be used to improve generation control over the structural and semantic composition of generated images.
\par
One such approach is Self-guidance for Controllable Image Generation \cite{epstein_diffusion_2023} (which we denote SGCIG to distinguish it from \cite{hu_self-guided_2023}). SGCIG is a zero-shot method designed to increase user control over structural and semantic elements of objects in images generated by text-to-image diffusion models. Incorporating similar ideas as \cite{hertz_prompt--prompt_2022}, the authors of SGCIG leverage representations from intermediate activations and attention maps to steer the generation process. SGCIG works by adding a series of guidance terms to the objective of the denoising network that each define a series of properties that can be used to perform image manipulations. Image edits can then be carried out by guiding properties to change in the pixel generation process. While the method is limited to the manipulation of objects explicitly stated in the conditioning text prompt, it represents a promising first step towards increased control over generated images. Diffusion Handles \cite{pandey_diffusion_2024} extend this to 3D object editing, using manipulated diffusion model activations to produce plausible edits.
\par
Depth-aware guidance (DAG) \cite{kim_depth-aware_2024} is a related method that uses semantic information from intermediate denoising network layers for improved depth-aware image synthesis. \citet{kim_depth-aware_2024} propose training depth predictors with limited depth-labeled data using internal U-Net backbone representations, similar to DDPM-Seg \cite{baranchuk_label-efficient_2022}. The used depth predictors are pixel-wise shallow MLP regressors estimating depth values from intermediate U-Net features $\mathbf{f}_t$ at timestep $t$. Features are concatenated across layers to form $\mathbf{g}_t$, with depth maps $\mathbf{d}_t = \text{MLP}(\mathbf{g}_t, t)$ generated using an appended time-embedding block. This depth predictor is trained using a limited depth-labeled dataset. To now guide the diffusion process toward depth-aware generation, two guidance strategies are introduced: Depth consistency guidance uses pseudo-labels with a consistency loss $\mathcal{L}_{\text{dc}}$ between weak and strong depth predictors, guiding the generation process using the gradient of $\mathcal{L}_{\text{dc}}$ with respect to $\mathbf{x}_t$ in a methodology similar to \cite{dhariwal2021diffusion_beat}. Depth prior guidance employs an additional small-resolution diffusion U-Net on the depth domain, adding noise to depth predictions and using a denoising objective $\mathcal{L}_{\text{dp}}$. The gradient of $\mathcal{L}_{\text{dp}}$ is treated like an external classifier gradient and added to the image generation objective. Combining both methods during training results in enhanced depth semantics in generated images.
\par
Perturbed Attention Guidance (PAG) \cite{ahn_self-rectifying_2024} is a sampling guidance method that improves generation quality for both conditional and unconditional settings. PAG does not require additional training or external pre-trained models. Instead, \citet{ahn_self-rectifying_2024} introduce an implicit discriminator $\mathcal{D}$ that differentiates between desirable and undesirable samples during the diffusion process, where $\mathbf{y}$ is a desirable and $\hat{\mathbf{y}}$ is an undesirable sample. The diffusion sampling process is then redefined to incorporate the derivative of the discriminator loss $\mathcal{L}_{\mathcal{D}}$. The score with undesirable label $\hat{\mathbf{y}}$ cannot be approximated using the existing denoising network $\epsilon_{\theta}(\mathbf{x}_t)$. Thus the score is estimated by perturbing the forward pass of a pre-trained denoising network, denoted by $\hat{\epsilon}_{\theta}$. PAG works by perturbing the self-attention maps in the diffusion U-Net, replacing them with an identity matrix to guide the sampling process away from degraded samples. The final noise prediction is obtained by feeding $\mathbf{x}_t$ into both $\epsilon_{\theta}(\cdot)$ and $\hat{\epsilon}_{\theta}(\cdot)$ to get the final noise prediction $\Tilde{\epsilon}_{\theta}$. PAG improves generation quality in both conditional and unconditional settings, and can be combined with existing guidance methods like classifier guidance.




\section{Challenges \& Future Directions}
\label{sec:challenges}

\subsection{General Challenges}

 Diffusion model-based representation learning is a novel research field with a lot of potential for theoretical and practical improvements. Improving synergies between representation learning and generative models is akin to a chicken-and-egg problem, where better diffusion models simultaneously lead to higher quality image representations, and better representation learning methods improve generative quality of diffusion models when applied to self-supervised guidance methods. Improved online-bootstrapping methods that provide guidance to diffusion models during training can be beneficial here.
\par
To conserve computation in diffusion models~\cite{luccioni2023power}, the sampling process has been significantly reduced to just a few steps~\cite{salimans_progressive_2022,salimans2024multistep} or even a single step~\cite{song2023consistency,heek2024multistep,luo2024diffinstruct}. However, maintaining the potential of representation learning with few sampling steps presents a challenge. 


\subsection{Potential Future Directions}

In many works discussed, the quality of representations learned by diffusion models is evaluated indirectly using task-specific metrics from auxiliary models. Interpretability and disentanglement are other important ways to evaluate representation efficacy and are currently underexplored. Methods enhancing the interpretability of the latent space can improve generation control and benefit a wide range of recognition tasks.  We look towards methods on interpretable direction discovery as have been proposed for GANs in \cite{voynov_unsupervised_2020} for inspiration, and see similar approaches for diffusion models as promising. While there are some recent works focusing on disentangled and interpretable representation learning in diffusion models (e.g., \cite{kwon_diffusion_2023, yue_exploring_2024,chefer_hidden_2024}), we feel that this area remains underserved.
\par
Current diffusion-based representation learning frameworks use U-Net and DiT backbones, which were primarily designed for generative tasks. Developing novel architectures tailored for representation learning is a promising area of research. Current transformer-based backbones are popular due to their scalability and performance, but their inability for parallel inference and the quadratic complexity of the attention mechanism are significant downsides, limiting their use for high-resolution images and long videos \cite{hu_zigma_2024}. Techniques like windowing \cite{liu_swin_2021}, sliding \cite{beltagy_longformer_2020}, and ring attention \cite{liu_ringattention_2024} help mitigate these issues, but complexity limitations remain. Recent works \cite{yan_diffusion_2024, fei_scalable_2024, hu_zigma_2024} have begun to utilize state-space diffusion models \cite{gu_mamba_2024, nguyen_s4nd_2022}, which offer linear complexity with respect to token sequence length, and are thus well suited to long token sequence modeling for both text \cite{mehta_long_2023} and images/video \cite{li_videomamba_2024, chen_video_2024}. The representation-learning capabilities of these models are yet to be fully analyzed, but we expect that conclusions drawn from diffusion models can also be applied to state-space models and their representation learning capabilities.
\par
We also see significant room for further research in using other generative models for representation learning. Flow Matching models \cite{lipman2023flow,liu2023rectified,albergo2022building} have recently gained prominence for their ability to maintain straight trajectories during generation. This characteristic results in faster inference, making Flow Matching a suitable alternative for addressing trajectory issues encountered in diffusion models. Their versatility has been demonstrated across various applications, including image~\cite{ taohu2023lfm,dao2023flowlatent}, video~\cite{video_fm}, depth~\cite{gui2024depthfm}, human motion~\cite{motionfm}, audio~\cite{le2023voicebox}, boosting diffusion models~\cite{Schusterbauer2023boosting,sauer2024fast,song2024flow}, and even text generation~\cite{flowseq}. The close relationship between Diffusion and Flow Matching models suggests that many of the diffusion representation learning frameworks can also be applied to Flow Matching models.

\ifCLASSOPTIONcompsoc
  \section*{Acknowledgments}
\else
  \section*{Acknowledgment}
\fi
We would like to thank Yuki Asano, Stefan Andreas Baumann, Timy Phan, and Frank Fundel for providing additional related literature.


\ifCLASSOPTIONcaptionsoff
  \newpage
\fi
{\small
\bibliographystyle{IEEEtranSN}
\bibliography{egbib}
}

\newpage

\begin{IEEEbiography}[{\includegraphics[width=1in,height=1.25in,clip,keepaspectratio]{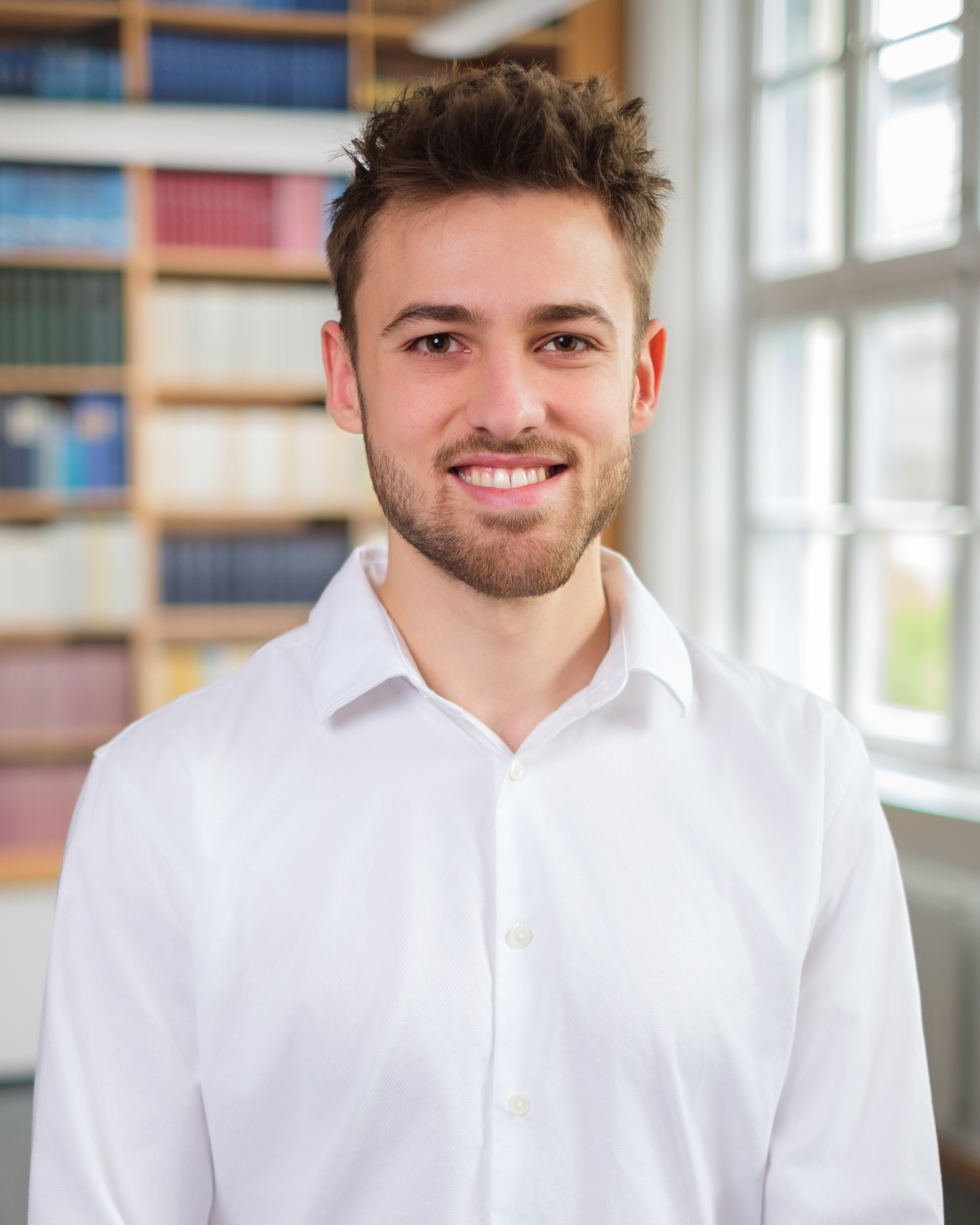}}]{Michael Fuest}
  is a research intern in the Computer Vision \& Learning Group at Ludwig-Maximilians-Universität München (LMU). He recently received his master's degree in Management \& Technology with a major in Computer Science from the Technical University of Munich, and is currently a visiting researcher at the MIT Laboratory for Information \& Decision Systems.
\end{IEEEbiography}

\begin{IEEEbiography}[{\includegraphics[width=1in,height=1.25in,clip,keepaspectratio]{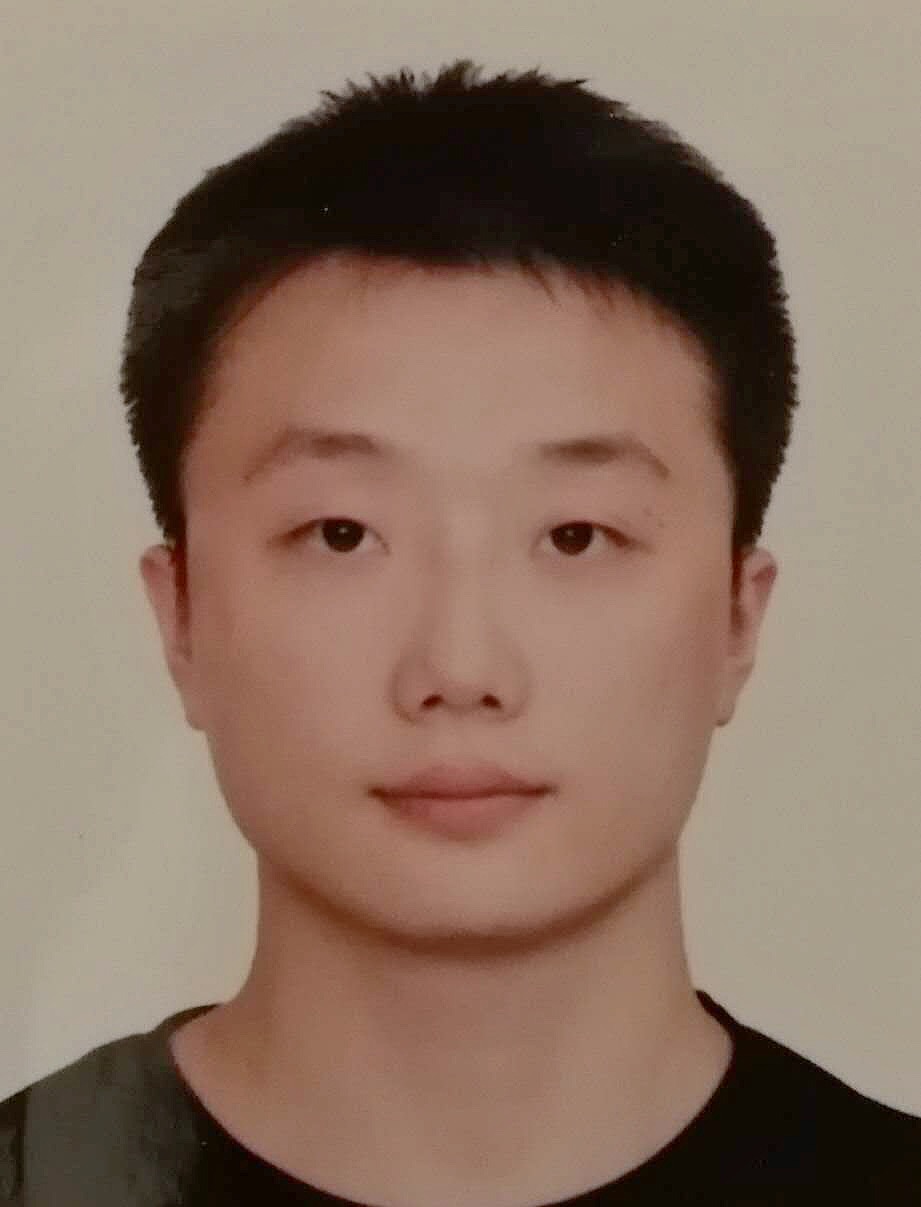}}]{Pingchuan Ma}
is a Ph.D. student in the Computer Vision \& Learning Group at Ludwig Maximilian University of Munich (LMU) and a Munich Center for Machine Learning (MCML) member. He previously received his master's degree in Applied Computer Science from Heidelberg University, where he developed an interest in deep metric learning and style transfer. He served as a reviewer for CVPR 2024 and NeurIPS 2024. His current research focuses on leveraging generative models for tasks beyond generation and exploring multi-modality representation learning.
\end{IEEEbiography}

\begin{IEEEbiography}[{\includegraphics[width=1in,height=1.25in,clip,keepaspectratio]{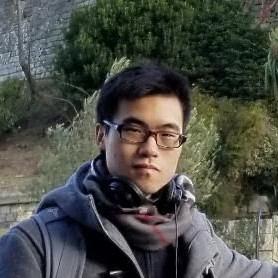}}]{Ming Gui}
is currently a PhD researcher at the Computer Vision \& Learning Group at Ludwig Maximilian University of Munich (LMU). He received his bachelor’s and master’s degree in electrical and computational engineering from the Technical University of Munich, where he developed interest in deep learning and computer vision. His research is presently centered around the development and enhancement of scalable generative models.
\end{IEEEbiography}

\begin{IEEEbiography}[{\includegraphics[width=1in,height=1.25in,clip,keepaspectratio]{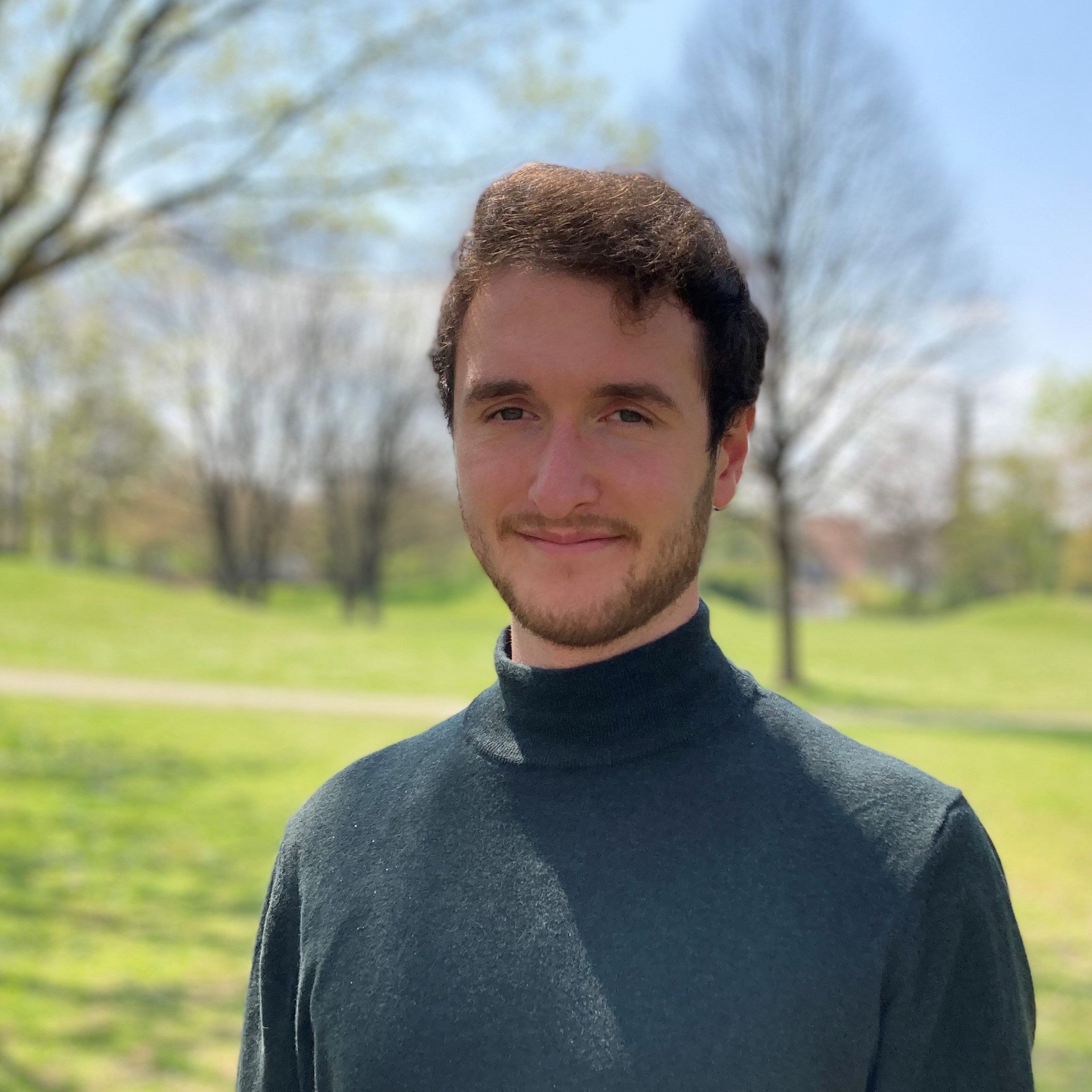}}]{Johannes Schusterbauer}
is a PhD Student at the University of Munich’s Computer Vision \& Learning Group. He did his undergraduate studies in Psychology and Computer Science, both at the University of Munich, followed by a Master’s degree in Intelligent Interactive Systems from Pompeu Fabra University in Barcelona. Besides works in the field of generative modeling, Johannes’ research focuses on the adaptability of Diffusion and Flow-based models for general image understanding.
\end{IEEEbiography}

\begin{IEEEbiography}
[{\includegraphics[width=1in,height=1.1in,clip,keepaspectratio]{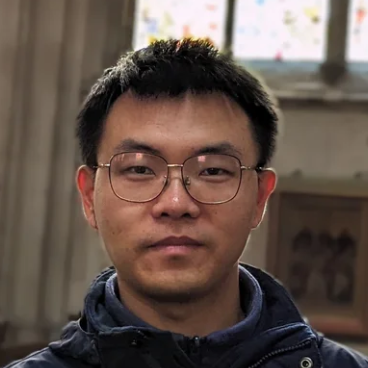}}]{Tao Hu}
 is a postdoctoral research fellow at Ludwig Maximilian University of Munich (LMU) where they are developing next generation of Stable Diffusion models. He obtained his Ph.D. degree in computer science from the University of Amsterdam in 2023 under the supervision of Cees Snoek. He was an intern at Megvii, Amazon AWS in 2017 and 2020. His Ph.D. research has been selected for the CVPR2023 and ICCV2023 Doctoral Consortium. He co-organized the ECCV 2024 Workshop on Audio-Visual Generative Learning. He has also served as the Area Chair for the CVPR AI4CC workshop and the CVPR 2024 Efficient Large Vision Model workshop. He has been a reviewer for top-tier computer vision and machine learning conferences for several years. His current research interests lie in scalable, flexible, and efficient generative models. 
\end{IEEEbiography}

\begin{IEEEbiography}[{\includegraphics[width=1in,height=1.25in,clip,keepaspectratio]{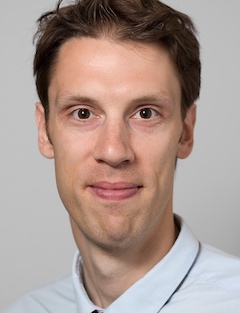}}]{Björn Ommer}
 is a professor at LMU and leads the Computer Vision \& Learning Group. He was previously with Heidelberg University's Department of Mathematics and Computer Science, IWR, and HCI.
He studied computer science and physics at the University of Bonn, completed his Ph.D. at ETH Zurich where his dissertation received the ETH Medal, and held a post-doc position with Jitendra Malik at UC Berkeley.
He is a member of the Bavarian AI Council, an editor for IEEE T-PAMI, an ELLIS Fellow, faculty of ELLIS unit Munich, a PI at the Munich Center for Machine Learning (MCML), and has held various roles at numerous CVPR, ICCV, ECCV, and NeurIPS conferences. He delivered the opening keynote at NeurIPS’23.
\end{IEEEbiography}

\end{document}